\def\year{2022}\relax
\documentclass[letterpaper]{article} 
\usepackage{aaai22}  
\usepackage{times}  
\usepackage{helvet}  
\usepackage{courier}  
\usepackage[hyphens]{url}  
\usepackage{graphicx} 
\usepackage{natbib}  
\usepackage{caption} 
\DeclareCaptionStyle{ruled}{labelfont=normalfont,labelsep=colon,strut=off} 
\usepackage{algorithm}
\usepackage{algorithmic}

\usepackage{amssymb}
\usepackage{amsmath}
\usepackage{arydshln}
\usepackage{booktabs}
\usepackage{hyperref}
\usepackage{subfig}
\usepackage{tikz}

\usepackage{newfloat}
\usepackage{listings}
\floatstyle{ruled}
\newfloat{listing}{tb}{lst}{}
\floatname{listing}{Listing}
\newcommand{\cellsize}{0.23}
\newcommand{\cellsizecm}{\cellsize cm}

\newcommand{\revised}[1]{{#1}}
\newcommand{\revisedmay}[1]{{#1}}

\newcommand{\notrudder}{A2C-RR}

\title{Adaptive Pairwise Weights for Temporal Credit Assignment}
\author{
    Zeyu Zheng\textsuperscript{\rm 1}\equalcontrib,
    Risto Vuorio\textsuperscript{\rm 2}\equalcontrib,
    Richard Lewis\textsuperscript{\rm 1},
    Satinder Singh\textsuperscript{\rm 1}
}
\affiliations{
    \textsuperscript{\rm 1}University of Michigan\\
    \textsuperscript{\rm 2}University of Oxford\\
    zeyu@umich.edu, risto.vuorio@cs.ox.ac.uk, rickl@umich.edu, baveja@umich.edu

}

\usepackage{bibentry}

\begin{document}

\maketitle

\begin{abstract}
How much credit (or blame) should an action taken in a state get for a future reward? This is the fundamental temporal credit assignment problem in Reinforcement Learning (RL). One of the earliest and still most widely used heuristics is to assign this credit based on a scalar coefficient, lambda (treated as a hyperparameter), raised to the power of the time interval between the state-action and the reward. In this empirical paper, we explore heuristics based on more general pairwise weightings that are functions of the state in which the action was taken, the state at the time of the reward, as well as the time interval between the two. Of course it isn't clear what these pairwise weight functions should be, and because they are too complex to be treated as hyperparameters we develop a metagradient procedure for learning these weight functions during the usual RL training of a policy. Our empirical work shows that it is often possible to learn these pairwise weight functions during learning of the policy to achieve better performance than competing approaches.
\end{abstract}

\section{Introduction
\label{sec:Introduction}}
The following \emph{umbrella problem}~\citep{osband2019behaviour} illustrates a fundamental challenge in most reinforcement learning (RL) problems, namely the \emph{temporal credit assignment} (TCA) problem. An RL agent takes an umbrella at the start of a cloudy morning and experiences a long day at work filled with various rewards uninfluenced by the umbrella, before needing the umbrella in the rain on the way home. The agent must learn to credit the take-umbrella action in the cloudy-morning state with the very delayed reward at the end of the day, while also learning to not credit the action with the many intervening rewards, despite their occurring much closer in time. More generally, the TCA problem is  how much credit or blame should an action taken in a state get for a future reward. 
One of the earliest and still most widely used heuristics for TCA comes from the celebrated TD($\lambda$)~\citep{sutton1988learning} family of algorithms, and assigns credit based on a scalar coefficient $\lambda$ raised to the power of the time interval between the state-action and the reward. Note that this is a recency and frequency heuristic, in that it assigns credit based on how recently and how frequently a state-action pair has occurred prior to the reward. It is important, however, to also note that this heuristic has not in any way shown to be the ``optimal'' way for TCA. In particular, in the umbrella problem the action of taking the umbrella on a cloudy morning will be assigned credit for the rewards achieved during the workday early on in learning and it is only after a lot of learning that this effect will diminish. Nevertheless, the recency and frequency heuristic has been adopted in most modern RL algorithms because it is so simple to implement, with just one hyperparameter, and because it has  been shown to allow for asymptotic convergence to the true value function under certain circumstances. 

In this empirical paper, we present two new families of algorithms for addressing TCA: one that generalises TD($\lambda$) and a second that generalises a Monte-Carlo algorithm. Specifically, our generalisation introduces pairwise weightings that are functions of  \emph{the state in which the action was taken}, \emph{the state at the time of the reward}, and \emph{the time interval between the two}. Of course, it isn't clear what this pairwise weight function should be, and it is too complex to be treated as a hyperparameter (in contrast to the scalar $\lambda$ in TD($\lambda$)). We develop a metagradient approach to learning the pairwise weight function at the same time as learning the policy of the agent. Like other metagradient algorithms, our algorithm has two loops: an outer loop that periodically updates the pairwise weight function in order to optimize the usual RL loss (policy gradient loss in our case) and an inner loop where the policy parameters are updated using the pairwise weight function set by the outer loop.

Our main contribution in this paper is a family of algorithms that contains within it the theoretically well understood TD($\lambda$) and Monte-Carlo algorithms. We show that the additional flexibility of our algorithms can yield benefit analytically in a simple illustrative example intended to build intuition and then empirically in more challenging TCA problems. A second contribution is the metagradient algorithm to learn such the pairwise-weighting function that parameterises our family of algorithms. Our empirical work is geared towards answering two questions: (1) Are the proposed pairwise weight functions able to outperform the best choice of $\lambda$ and other baselines? (2) Is our metagradient algorithm able to learn the pairwise weight functions fast enough to be worth the extra complexity they introduce? 

\section{Related Work} 
\label{sec:related}
\revised{
Several heuristic methods have been proposed to address the long-term credit assignment problem in RL. 
RUDDER~\citep{arjona2019rudder} trains a LSTM~\citep{hochreiter1997long} to predict the return of an episode given the entire state and action sequence and then conducts contribution analysis with the LSTM to redistribute rewards to state-action pairs. 
Synthetic Returns (SR)~\citep{raposo2021synthetic} directly learns the association between past events and future rewards and use it as a proxy for credit assignment. 
Different from the predictive approach of RUDDER and SR, Temporal Value Transport (TVT)~\citep{hung2019optimizing} augments the agent with an external memory module and utilizes the memory retrieval as a proxy for transporting future value back to related state-action pairs. 
We compare against TVT by using their published code, and we take inspiration from the core reward-redistribution idea from RUDDER and implement it within our policy gradient agent as a comparison baseline (because the available RUDDER code is not directly applicable). We do not compare to SR because their source code is not available. 

We also compare against two other algorithms that are more closely related to ours in their use of metagradients. Xu et al.~\citep{xu2018meta} adapt $\lambda$ via metagradients rather than tuning it via hyperparameter search, thereby improving over the use of a fixed-$\lambda$ algorithm. 
The Return Generating Model (RGM)~\citep{wang2019beyond} generalizes the notion of return from exponentially discounted sum of rewards to a more flexibly weighted sum of rewards where the weights are adapted via metagradients during policy learning. RGM takes the entire episode as input and generates one weight for each time step. In contrast, we study pairwise weights as explained below.

Some recent works address counterfactual credit assignment where classic RL algorithms struggle~\citep{harutyunyan2019hindsight,mesnard2020counterfactual,van2020expected}. Although they are related to our work in that they also address the TCA problem, we do no compare to them because our work does not focus on the counterfactual aspect.
}

\section{Pairwise Weights for Advantages}
\label{sec:pairwise}
At the core of our contribution are new parameterizations of functions for computing advantages used in policy gradient methods. 
Therefore, we briefly review advantages in policy gradient methods and TD$(\lambda)$ as our points of departure.

\paragraph{Background on Advantages, Policy Gradient, and TD$(\lambda)$.}
We assume an episodic RL setting. 
The agent's policy $\pi_{\theta}$, parameterized by $\theta$, maps a state $S$ to a probability distribution over the actions. 
Within each episode, at time step $t$, the agent observes the current state $S_{t}$, takes an action $A_{t} \sim \pi_{\theta}(\cdot | S_{t})$, and receives the reward $R_{t+1}$.
The return is denoted by $G_{t} = \sum_{k=t+1}^{T} \gamma^{k-t-1} R_{k}$ where $\gamma$ is the discount factor and $T$ denotes the length of the episode. 
The state-value function $V^{\pi}$ is defined as 
\begin{equation}
V^{\pi}(s) = \mathbb{E}_{\pi} [G_{t} | S_{t} = s],
\end{equation}
and the action-value function $Q^{\pi}$ is defined as 
\begin{equation}
Q^{\pi}(s, a) = \mathbb{E}_{\pi} [G_{t} | S_{t} = s, A_{t} = a].
\end{equation}
The notation $\mathbb{E}_{\pi}[\cdot]$ denotes the expected value of a random variable given that the agent follows the policy $\pi$. Because the policy is parameterized by $\theta$, we will use $\mathbb{E}_{\pi}[\cdot]$ and $\mathbb{E}_{\theta}[\cdot]$ interchangeably. The advantage function is defined as
\begin{equation}
\Psi^{\pi}(s, a) = Q^{\pi}(s, a) - V^{\pi}(s, a).
\end{equation}
For brevity, we will omit the superscript $\pi$ on $V$, $Q$, and $\Psi$.

The performance measure for the policy $\pi_{\theta}$, denoted by $J(\theta)$, is defined as the expected sum of the rewards when the agent behaves according to $\pi_{\theta}$, i.e.,
\begin{equation}
J(\theta) = \mathbb{E}_{\theta} [\sum_{t=1}^{T} \gamma^{t-1} R_{t}],
\end{equation}
The gradient of $J(\theta)$ w.r.t the policy parameters $\theta$ is~\citep{sutton2000policy,williams1992simple}
\begin{equation}\label{eq:policy-gradient}
\nabla_{\theta} J(\theta) = \mathbb{E}_{\theta} \Big[\big(G_{t} - b(S_t)\big) \nabla_{\theta}\log \pi_{\theta}(A_{t} | S_{t})\Big],
\end{equation}
where $b(S_t)$ is a baseline function for variance reduction. If we choose the state-value function $V$ as the baseline function, then Eq.~\ref{eq:policy-gradient} can be rewritten as~\citep{schulman2015high}
\begin{equation}
\nabla_{\theta} J(\theta) = \mathbb{E}_{\theta} \Big[\Psi(S_{t}, A_{t}) \nabla_{\theta}\log \pi_{\theta}(A_{t} | S_{t})\Big].
\end{equation}

Since the true state-value function $V$ is usually unknown, an approximation $v$ is used in place of $V$, which leads to a Monte-Carlo (MC) estimation of $\Psi$:
\begin{equation}
\hat{\Psi}^{\text{MC}}_{t} = G_{t} - v(S_{t}).
\end{equation}
However, $\hat{\Psi}^{\text{MC}}$ usually suffers from high variance.
To reduce variance, the approximated state-value function $v$ is used to estimate the return as in the TD($\lambda$) algorithm using the eligibility trace parameter $\lambda$; specifically the new form of the return, called $\lambda$-return is a weighted sum of $n$-step truncated corrected returns where the correction uses the estimated value function after $n$-steps. 
The corresponding $\lambda$-estimator is (see~\citep{schulman2015high} for a full derivation)
\begin{equation}
\hat{\Psi}^{(\lambda)}_{t} = \sum_{k=t+1}^{T} (\gamma\lambda)^{k-t-1} \delta_{k},
\end{equation}
where $\delta_{t} = R_{t} + \gamma v(S_{t}) - v(S_{t-1})$ is the TD-error at time $t$. 
As a special case, when $\lambda=1$, it recovers the MC estimator~\citep{schulman2015high}.
As noted above, the value for $\lambda$ is usually manually tuned as a hyperparameter. Adjusting $\lambda$ provides a way to tradeoff bias and variance in $\hat{\Psi}^{(\lambda)}$ (this is absent in $\hat{\Psi}^{\text{MC}}$).  Below we present two new estimators that are analogous in this regard to $\hat{\Psi}^{(\lambda)}$ and $\hat{\Psi}^{\text{MC}}$.

\paragraph{Proposed Heuristic 1: Advantages via Pairwise Weighted Sum of TD-errors.}
Our first new estimator, denoted PWTD for {\bf P}airwise {\bf W}eighted {\bf TD}-error, is a strict generalization of the $\lambda$-estimator above and is defined as follows:
\begin{equation}
    \hat{\Psi}^{\text{PWTD}}_{\eta, t} = \sum_{k=t+1}^{T} f_{\eta}(S_{t}, S_{k}, k - t) \delta_{k},
\end{equation}
where $f_{\eta}(S_{t}, S_{k}, k - t) \in [0, 1]$, parameterized by $\eta$, is the scalar weight given to the TD-error $\delta_{k}$ as a function of the state to which credit is being assigned, the state at which the TD-error is obtained, and the time interval between the two. 
Note that if we choose $f(S_{t}, S_{k}, k - t) = (\gamma\lambda)^{k - t - 1}$, it recovers the usual $\lambda$-estimator $\hat{\Psi}^{(\lambda)}$.

\paragraph{Proposed Heuristic 2: Advantages via Pairwise Weighted Sum of Rewards.}
Instead of generalizing from the $\lambda$-estimator, we can also generalize from the MC estimator via pairwise weighting. Specifically, the new pairwise-weighted return is defined as
\begin{equation} \label{eq:pw-return}
    G^{\text{PWR}}_{\eta,t} = \sum_{k=t+1}^{T} f_{\eta}(S_{t}, S_{k}, k - t) R_{k},
\end{equation}
where $f_{\eta}(S_{t}, S_{k}, k - t) \in [0, 1]$ is the scalar weight given to the reward $R_{k}$.
The corresponding advantage estimator, denoted PWR for {\bf P}airwise {\bf W}eighted {\bf R}eward, then is:
\begin{equation} \label{eq:pw-advantage}
    \hat{\Psi}^{\text{PWR}}_{\eta, t} = G^{\text{PWR}}_{\eta,t} - v^{\text{PWR}}(S_{t}),
\end{equation}
where $V^{\text{PWR}}(s) = \mathbb{E}_{\theta} [G^{\text{PWR}}_{\eta,t} | S_{t} = s]$ and $v^{\text{PWR}}$ is an approximation of $V^{\text{PWR}}$. 
Note that if we choose $f(S_{t}, S_{k}, k - t) = \gamma^{k - t - 1}$, we can recover the MC estimator $\hat{\Psi}^{\text{MC}}$.

\revised{
The benefit of the additional flexibility provided by these new estimators highly depends on the choice of the pairwise weight function $f$. 
As we will demonstrate in the simple example below, the new estimators can yield lower variance and benefit policy learning if the function $f$ captures the underlying credit assignment structure of the problem. 
On the other hand, the new estimators may not even be well-defined in the infinite-horizon setting if the pairwise weight function is chosen wrongly because the weighted sum of TD-errors/rewards could be unbounded. 
Designing a good pairwise weight function by hand is challenging because it requires both domain knowledge to capture the credit assignment structure and careful tuning to avoid harmful consequences. 
Thus we propose a metagradient algorithm to \emph{learn} the pairwise weight function such that it benefits policy learning, as detailed in \S~\ref{sec:method}.
}

\paragraph{An Illustrative Analysis of the Benefit of the PWR Estimator.} 
\begin{figure}
\centering
\includegraphics[width=0.7\linewidth]{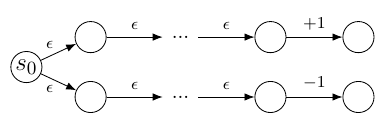}
    \caption{A simple illustrative MDP. The initial action determines the final reward but does not impact the intermediate rewards. The consequence of the initial action is delayed.}
    \label{fig:umbrella}
\end{figure}
Consider the simple-MDP version of the umbrella problem in Figure~\ref{fig:umbrella}. 
Each episode starts at the leftmost state, $s_{0}$, and consists of $T$ transitions. The only choice of action is at $s_0$ and it determines the reward on the last transition. A noisy reward $\epsilon$ is sampled for each intermediate transition independently from a distribution with mean $\mu$ and variance $\sigma^2 > 0$. These intermediate rewards are independent of the initial action. 
We consider the undiscounted setting in this example. The expected return for state $s_{0}$ under policy $\pi$ is 
\[
V(s_{0}) = \mathbb{E}_{\pi}[G_{0}] = (T - 1) \mu + \mathbb{E}_{\pi} [R_{T}].
\]
For any initial action $a_{0}$, the advantage is
\begin{align*}
\Psi(s_{0}, a_{0}) &= \mathbb{E}_{\pi} [ G_{0} | a_{0} ] - V(s_{0}) = \mathbb{E}_{\pi}[R_{T} | a_{0}] - \mathbb{E}_{\pi} [R_{T}].
\end{align*}
Consider pairwise weights for computing $\hat{\Psi}^{\text{PWR}}(s_{0}, a_{0})$ that place weight only on the final transition, and zero weight on the noisy intermediate rewards, capturing the notion that the intermediate rewards are not influenced by the initial action choice.
More specifically, we choose $f$ such that for any episode, $w_{0T} = 1$ and $w_{ij} = 0$ for other $i$ and $j$.
The shorthand $w_{ij}$ denotes $f(S_{i}, S_{j}, j - i)$ for brevity. The expected parameterized reward sum for the initial state $s_{0}$ is
\begin{align*}
V^{\text{PWR}}(s_{0}) &= \mathbb{E}_{\pi}[G_{\eta,0}] = \mathbb{E}_{\pi}[\sum_{i=t}^{T} w_{0t} R_{t}] = \mathbb{E}_{\pi} [R_{T}].
\end{align*}
If $v^{\text{PWR}}$ is correct, for any initial action $a_{0}$, the pairwise-weighted advantage is the same as the regular advantage: 
\begin{align*}
    \mathbb{E}_{\pi} [\hat{\Psi}^{\text{PWR}}_{\eta}(s_{0}, a_{0})] &= \mathbb{E}_{\pi} [ G_{\eta,0} - v^{\text{PWR}}(s_{0}) | a_{0} ] \\
    &= \mathbb{E}_{\pi} [ \sum_{t=1}^{T} w_{0t} R_{t} ] - V^{\text{PWR}}(s_{0}) \\
    &= \mathbb{E}[R_{T} | a_{0}] - \mathbb{E}_{\pi} [R_{T}] 
    = \Psi(s_{0}, a_{0}).
\end{align*}
As for variance, for any initial action $a_{0}$, $[G_{\eta,0} | a_0]$ is deterministic because of the zero weight on all the intermediate rewards and thus $\hat{\Psi}^{\text{PWR}}_{\eta}(s_0,a_0)$ has zero variance. The variance of $\hat{\Psi}^{\text{MC}}(s_{0}, a_{0})$ on the other hand is $(T - 1) \sigma^2 > 0$.
Thus, in this illustrative example $\hat{\Psi}^{\text{PWR}}$ yields an unbiased advantage estimator with far lower variance than $\hat{\Psi}^{\text{MC}}$.

Our example exploited knowledge of the domain to set weights that would yield an unbiased advantage estimator with reduced variance, thereby providing some intuition on how a more flexible return might in principle yield benefits for learning. 
Of course, in general RL problems will have the umbrella problem in them to varying degrees.  But how can these weights be set by the agent itself, without prior knowledge of the domain? We turn to this question next.

\section{A Metagradient Algorithm for Adapting Pairwise Weights}
\label{sec:method}

Recently metagradient methods have been developed to learn various kinds of parameters that would otherwise be set by hand or by manual hyperparameter search; these include discount factors~\citep{xu2018meta,zahavy2020self}, intrinsic rewards~\citep{zheng2018learning,rajendran2019should,zheng2020can}, auxiliary tasks~\citep{veeriah2019discovery}, constructing general return functions~\citep{wang2019beyond}, and discovering new RL objectives~\citep{oh2020discovering,xu2020meta}. 
We use the metagradient algorithm from \cite{xu2018meta} for training the pairwise weights. The algorithm consists of an outer loop learner for the pairwise weight function, which is driven by a conventional policy gradient loss and an inner loop learner driven by a policy-gradient loss based on the new pairwise-weighted advantages.
An overview of the algorithm is in the appendix.
For brevity, we use $\hat{\Psi}_{\eta}$ to denote $\hat{\Psi}^{\text{PWTD}}_{\eta}$ or $\hat{\Psi}^{\text{PWR}}_{\eta}$ unless it causes ambiguity.

\paragraph{Learning in the Inner Loop.}
In the inner loop, the pairwise-weighted advantage $\hat{\Psi}_{\eta}$ is used to compute the policy gradient. 
We rewrite the gradient update from Eq.~\ref{eq:policy-gradient} with the new advantage as
\begin{equation}
\nabla_{\theta} J_{\eta}(\theta) = \mathbb{E}_{\tau \sim \pi_{\theta}} [\sum_{t=0}^{T-1} \hat{\Psi}_{\eta, t} \nabla_{\theta}\log \pi_{\theta}(A_{t}|S_{t}) ],
\end{equation}
where $\tau$ is a trajectory sampled by executing $\pi_{\theta}$. 
The overall update to $\theta$ is
\begin{equation} \label{eq:theta-update}
\nabla_{\theta}J^{\text{inner}}(\theta) = \nabla_{\theta}J_{\eta}(\theta) + \beta^{\mathcal{H}} \nabla_{\theta} \mathcal{H}(\pi_{\theta}),
\end{equation}
where $\mathcal{H}(\theta)$ is the usual entropy regularization term~\citep{mnih2016asynchronous} and $\beta^{\mathcal{H}}$ is a mixing coefficient. We apply gradient ascent to update the policy parameters and the updated parameters are denoted by $\theta'$.

Computing $\hat{\Psi}^{\text{PWR}}_{\eta}$ with Equation~\ref{eq:pw-advantage} requires a value function predicting the expected pairwise-weighted sums of rewards. 
We train the value function, $v_{\psi}$ with parameters $\psi$, along with the policy by minimizing the mean squared error between its output $v_{\psi}(S_{t})$ and the pairwise-weighted sum of rewards $G_{\eta,t}$.
The objective for training $v_{\psi}$ is
\begin{equation}\label{eq:value-loss}
J^{v}_{\eta}(\psi) = \mathbb{E}_{\tau \sim \pi_{\theta}} [\sum_{t=0}^{T-1} (G_{\eta,t} - v_{\psi}(S_{t}))^{2}].
\end{equation}
Note that $\hat{\Psi}^{\text{PWTD}}_{\eta}$ does not need this extra value function.

\paragraph{Updating $\eta$ via Metagradient in the Outer Loop.} 
To update $\eta$, the parameters of the pairwise weight functions, we need to compute the gradient of the usual policy loss w.r.t. $\eta$ through the effect of $\eta$ on the inner loop's updates to $\theta$. 
\begin{equation}\label{eq:eta-update}
    \nabla_{\eta}J^{\text{outer}}(\eta) = \nabla_{\theta'}J(\theta') \nabla_{\eta}\theta'.
\end{equation}
where, 
\begin{equation}
\nabla_{\theta'}J(\theta') = \mathbb{E}_{\tau' \sim \pi_{\theta'}} [\sum_{i=0}^{T-1} \Psi_{t} \nabla_{\theta'} \log \pi_{\theta'}(A_{t} | S_{t})],
\end{equation}
$\tau'$ is another trajectory sampled by executing the updated policy $\pi_{\theta'}$ and $\Psi_{t}$ is the regular advantage. 

Note that we need two trajectories, $\tau$ and $\tau'$, to make one update to the meta-parameters $\eta$. 
The policy parameters $\theta$ are updated after collecting trajectory $\tau$. 
The next trajectory $\tau'$ is collected using the updated parameters $\theta'$. 
The $\eta$-parameters are updated  on $\tau'$. 
In order to make more efficient use of the data, we follow~\citep{xu2018meta} and reuse the second trajectory $\tau'$ in the next iteration as the trajectory for updating $\theta$. 
In practice we use modern auto-differentiation tools to compute Equation~\ref{eq:eta-update} without applying the chain rule explicitly. 
Computing the regular advantage requires a value function for the regular return.
This value function is parameterized by $\phi$ and updated to minimize the squared loss analogously to  $v_\phi$.

\section{Experiments}
\label{sec:experiments}

We present three sets of experiments. 
The first set (\S\ref{subsec:resetting}) uses simple tabular MDPs  that allow visualization of the pairwise weights learned by Meta-PWTD and -PWR. The results show that the metagradient adaptation both \emph{increases} and \emph{decreases} weights in a way that can be interpreted as reflecting explicit credit assignment and variance reduction. 
In the second set (\S\ref{subsec:key-to-door}) we test Meta-PWTD and -PWR with neural networks in the benchmark credit assignment task \emph{Key-to-Door}~\citep{hung2019optimizing}. We show that Meta-PWTD and -PWR outperform several existing methods for directly addressing credit assignment, as well as TD($\lambda$) methods, and show again that the learned weights reflect domain structure in a sensible way. 
In the third set (\S\ref{subsec:bsuite}), we evaluate Meta-PWTD and -PWR in two benchmark RL domains, \texttt{bsuite}~\citep{osband2019behaviour} and Atari, and show that our methods do not hinder learning when environments do not pose idealized long-term credit assignment challenges.

\begin{figure}
\centering
{\includegraphics[width=0.45\linewidth,trim=5 30 5 30]{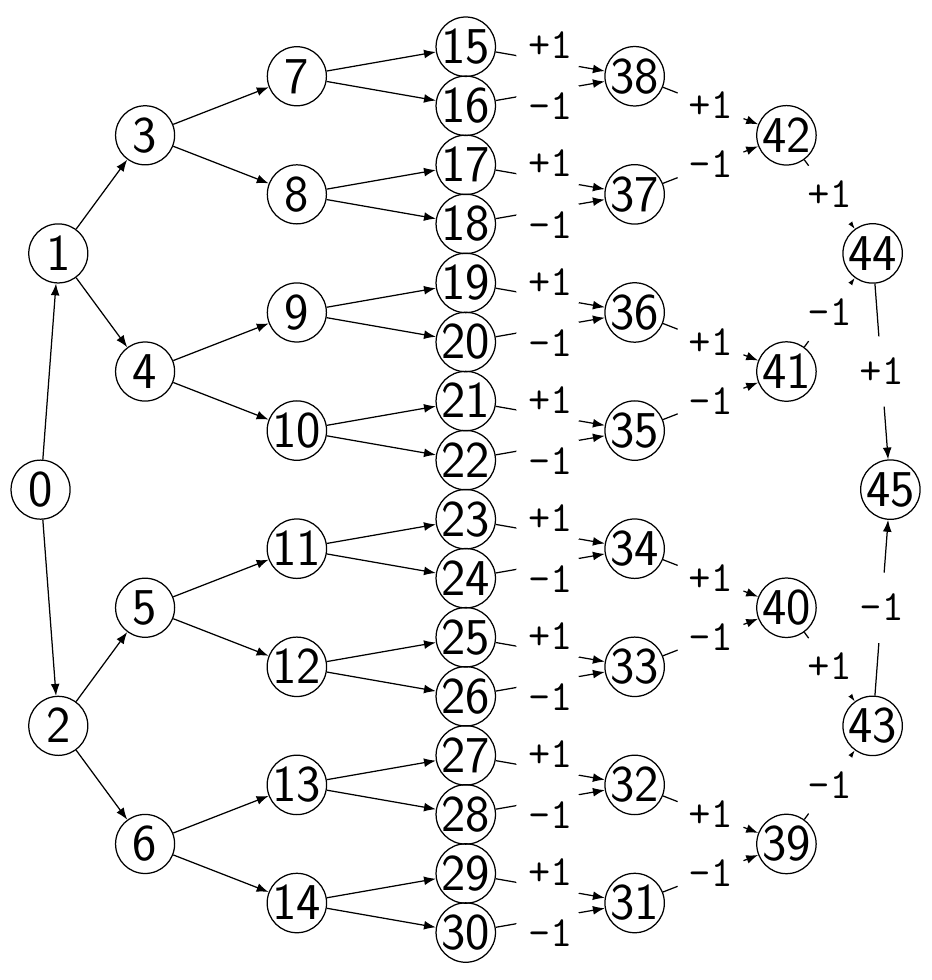}}\hspace{1.5em}
{\includegraphics[width=0.45\linewidth,trim=5 30 5 30]{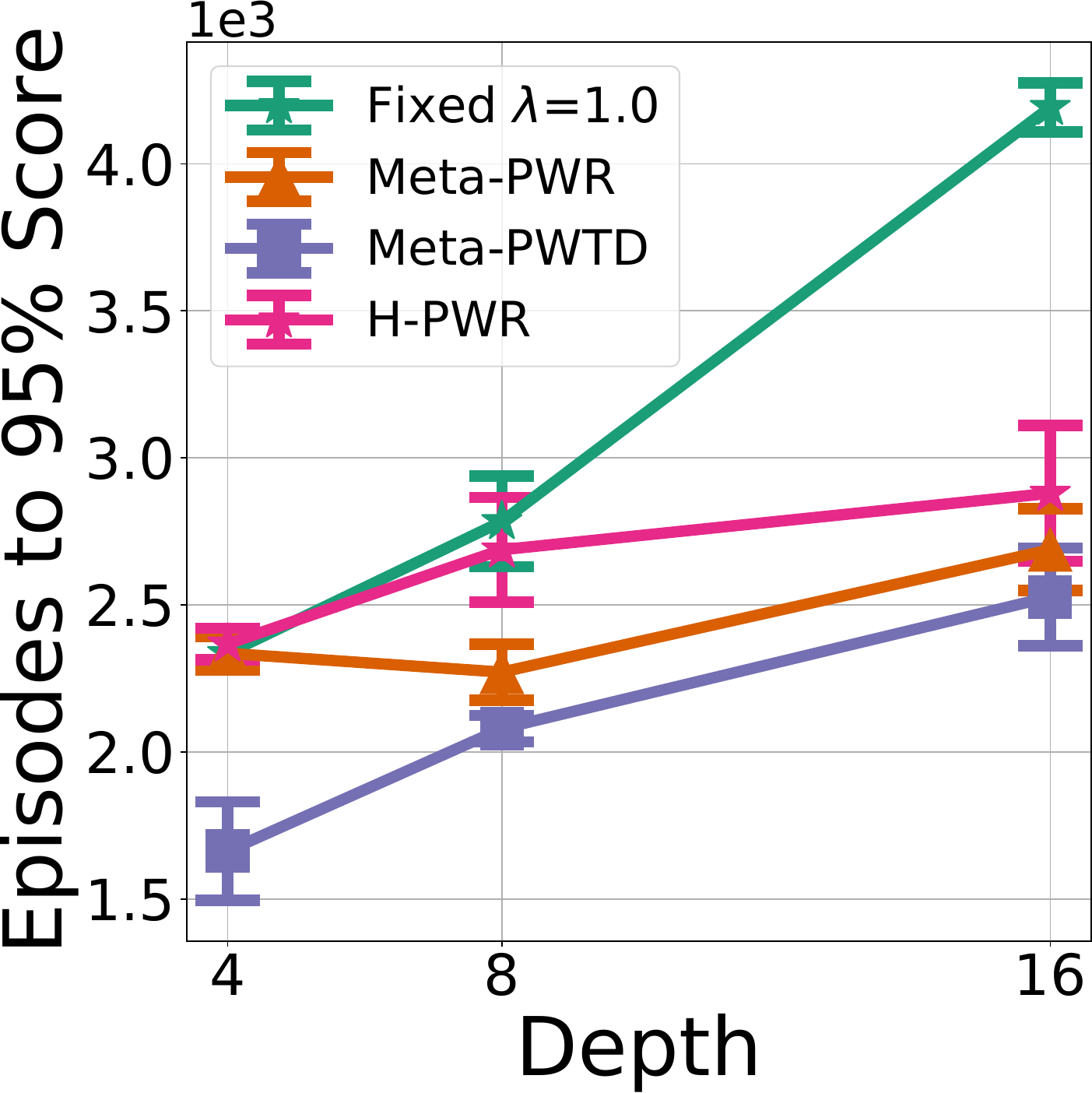}}
\caption{
(Left) Depth $8$ DAG environment with choice of two actions at each state and rewards along transitions.
(Right) Learning performance of regular return, handcrafted weights, and fixed meta-learned weights. Results are averaged over $5$ independent runs. Low is good.}
\label{fig:dag}
\end{figure}

\subsection{Learned Pairwise Weights in A Simple MDP}
\label{subsec:resetting}
Consider the environment represented as a DAG in Figure~\ref{fig:dag} (left). 
In each state in the left part of the DAG (states $0$--$14$, the \emph{first phase}), the agent chooses one of two actions but receives no reward.
In the remaining states (states $15$--$44$, the \emph{second phase}) the agent has only one action available and it receives a reward of $+1$ or $-1$ at each transition. Crucially, the rewards the agent obtains in the second phase are a consequence of the action choices in the first phase because they determine which states are encountered in the second phase.  There is an interesting credit assignment problem with a nested structure; for example, the action chosen at state $1$ determines the reward received later upon transition into state $44$.
We refer to this environment as the \emph{Depth $8$ DAG} and also report results below for depths $4$ and $16$.

In the DAG environments we use a tabular policy, value function, and meta-parameter representations.
The parameters $\theta$, $\psi$, $\phi$, and $\eta$ represent the policy, baseline for the weighted return, baseline for the regular return, and meta-parameters respectively.
The $\eta$ parameters are a $|S| \times |S|$ matrix. The entry on the $i$th row and the $j$th column defines the pairwise weight for computing the contribution of reward at state $j$ to the return at state $i$.
A sigmoid is used to bound the weights to $[0, 1]$, and the $\eta$ parameters are initialized so that the pairwise weights are close to $0.5$.
\revisedmay{Note that even in a tabular domain such as the DAG, setting the credit assignment weights by random search would be infeasible due to the number of possible weight combinations. This problem is exacerbated by larger domains discussed in the following sections. For this reason, the metagradient algorithm is a promising candidate for setting the weights.}

\paragraph{Visualizing the Learned Weights via Inner-loop Reset.}

\begin{figure}
\centering \rotatebox{90}{\hspace{1em} Handcrafted}\hspace{0.3em}\includegraphics[width=0.8\linewidth]{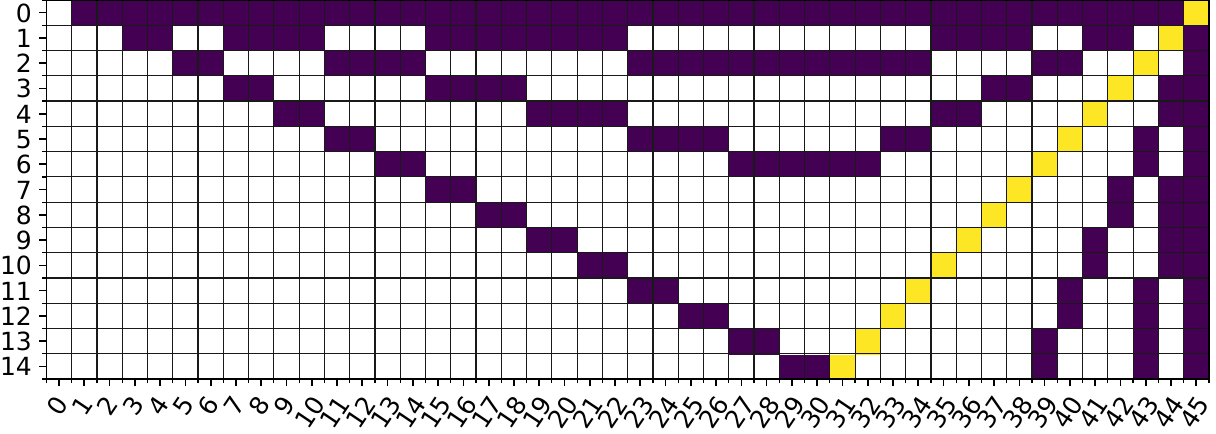} \\
\centering \rotatebox{90}{\hspace{1.1em} Meta-PWR}\hspace{0.3em}\includegraphics[width=0.8\linewidth]{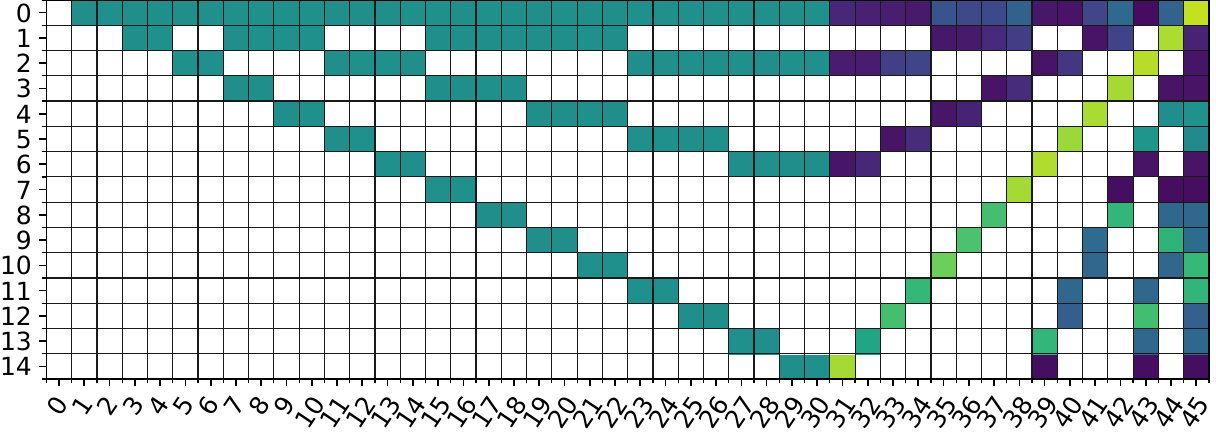} \\
\centering \rotatebox{90}{\hspace{1em} Meta-PWTD}\hspace{0.3em}\includegraphics[width=0.8\linewidth]{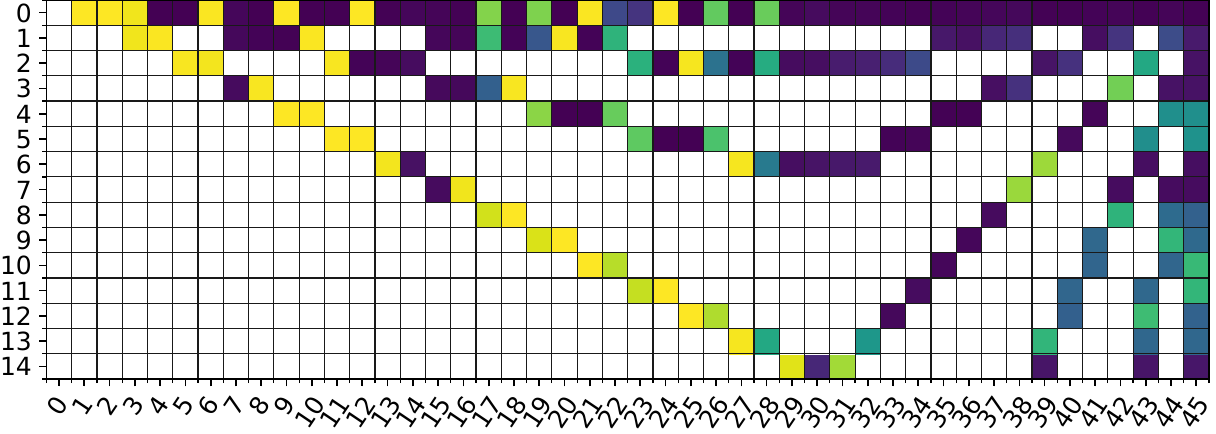} \\
\centering {\includegraphics[width=0.8\linewidth, trim=-17 0 8 0]{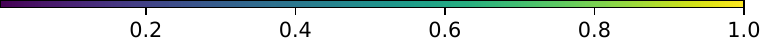}}
\caption{Inner loop-reset weight visualization: Top: Handcrafted pairwise weights for Depth 8 DAG; rows and columns correspond to states in Fig.~\ref{fig:dag}. Middle: Meta-learned weights for Depth $8$ DAG for Meta-PWR and Bottom: Meta-PWTD.}
\label{fig:dag_weights}
\end{figure}

To clearly see the most effective weights that metagradient learned for a random policy, we repeatedly reset the policy parameters to a random initialization while continuing to train the meta-parameters until convergence.
More specifically: the meta-parameters $\eta$ are trained repeatedly by randomly initializing $\theta$, $\psi$, and $\phi$ and running the inner loop for $16$ updates for each outer loop update.
Following existing work in metagradient~\citep{veeriah2019discovery,zheng2020can}, the outer loop objective is evaluated on all $16$ trajectories sampled with the updated policies.
The gradient of the outer loop objective on the $i\text{th}$ trajectory with respect to $\eta$ is backpropagated through all of the preceding updates to $\theta$.
Hyperparameters are provided in the appendix.

What pairwise weights would accelerate learning in this domain? 
Figure~\ref{fig:dag} (top) visualizes a set of \emph{handcrafted} weights for $\hat{\Psi}^{\text{PWR}}$ in the Depth $8$ DAG; each row in the grid represents the state in which an action advantage is estimated, and each column the state in which a future reward is experienced. For each state pair $(s_i, s_j)$ the weight is $1$ (yellow) only if the reward at $s_j$ depends on the action choice at $s_i$, else it is zero (dark purple; the white pairs are unreachable). 
Figure~\ref{fig:dag} (middle) shows the corresponding weights learned by Meta-PWR. Importantly, the learned pairwise weights have been \emph{increased} for those state pairs in which the handcrafted weights are $1$ and have been \emph{decreased} for those state pairs in which the handcrafted weights are $0$. As in the analysis of the simple domain in \S\ref{sec:pairwise}, these weights will result in lower variance advantage estimates.

The same reset-training procedure was applied to $\Psi^{\text{PWTD}}$. 
Figure~\ref{fig:dag} (bottom) visualizes the resulting weights. Since the TD-errors depend on the value function which are nonstationary during agent learning, we expect different weights to emerge at different points in training; the presented weights are but one snapshot. 
But a clear contrast to reward weighting can be seen: high weights are placed on transitions in the first phase of the DAG, which yield no rewards---because the TD-errors at these transitions do provide signal once the value function begins to be learned. 
In the appendix, we explicitly probe the adaptation of  $\Psi^{\text{PWTD}}$ to different points in learning by  modifying the value function in reset experiments, and show that the weights indeed adapt sensibly to differences in the accuracy of the value function.

\paragraph{Evaluation of the Learned Pairwise Weights.}
After the $\theta$-reset training of the pairwise-weights completed, we used them to train a new set of $\theta$ parameters, fixing the pairwise weights during learning. 
Figure~\ref{fig:dag} (right) shows the number of episodes to reach $95\%$ of the maximum score in each DAG, for policies trained with regular returns, handcrafted weights (H-PWR), and meta-learned weights. Using the meta-learned weights learned as fast as (indeed faster than) using the handcrafted weights, and both were faster than the regular returns, with the gap increasing for larger DAG-depth. 
We conjecture that the learned weights performed even better than the handcrafted weights because the learned weights adapted to the dynamics of the inner-loop policy learning procedure whereas the handcrafted weights did not.
Learning curves in the appendix show that all method achieved the optimal performance in the end.

\begin{figure}
    \centering
    \begin{minipage}[t]{0.99\textwidth}
      \input{figures/key_room}
      \input{figures/apple_room}
      \input{figures/door_room}
    \end{minipage}
    {\includegraphics[height=0.15\textwidth]{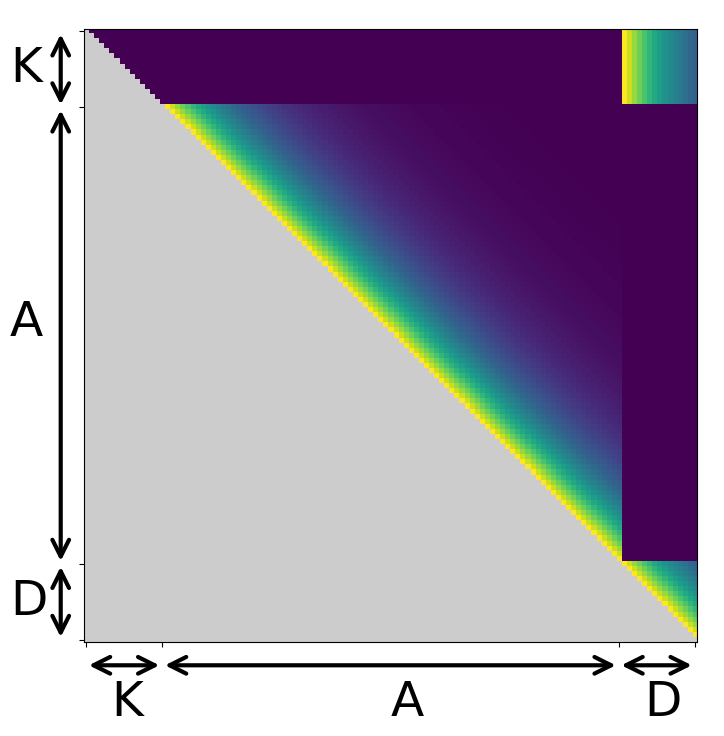}}~~~~{\includegraphics[height=0.15\textwidth]{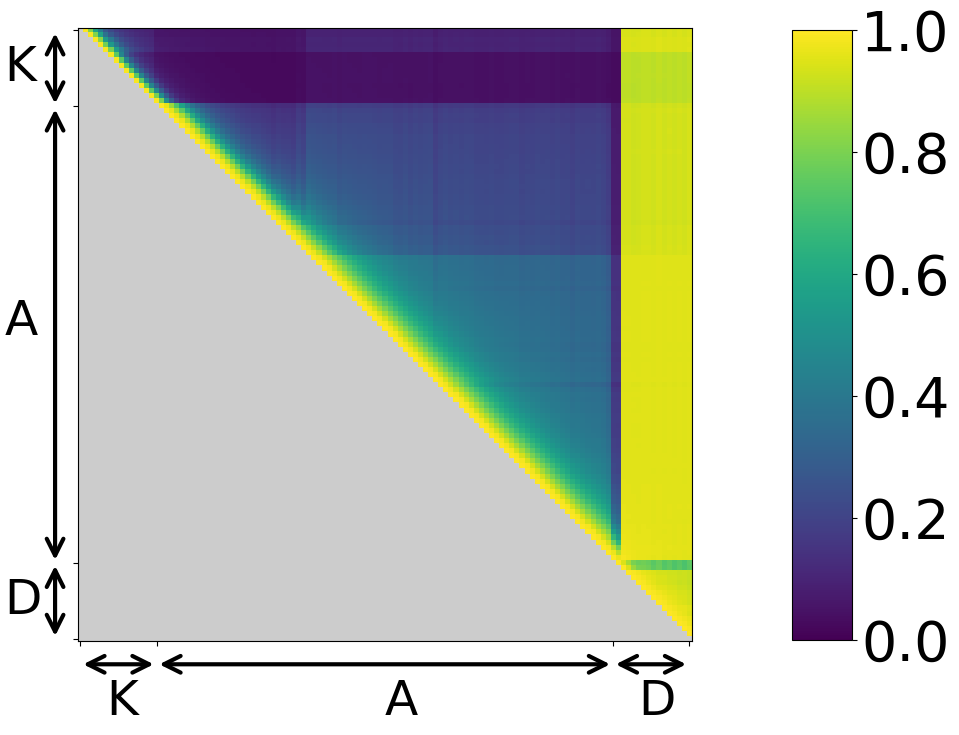}}
    \caption{
    (Top) The three phases in KtD. The blue circle denotes the agent.
    (Bottom left) Visualization of Handcrafted weights in the KtD experiment.
    (Bottom right) Weights learned by Meta-PWR in the $\mu=5, \sigma=5$ configuration.
    }
    \label{fig:ktd-layout}
\end{figure}

\subsection{The Key-to-Door Experiments}
\label{subsec:key-to-door}

We evaluated Meta-PWTD and -PWR the Key-to-Door (KtD) environment~\citep{hung2019optimizing} that is an elaborate umbrella problem that was designed to show-off the TVT algorithm's ability to solve TCA. We varied properties of the domain to vary the credit assignment challenge. We compared the learning performance of our algorithms to a version of $\hat{\Psi}^{\text{PWR}}$ that uses fixed handcrafted pairwise weights and no metagradient adaptation, as well as to the following \textbf{five} baselines (see related work in \S \ref{sec:related}): 
\textbf{(a)} best fixed-$\lambda$: Actor-Critic (A2C)~\citep{mnih2016asynchronous} with a best fixed $\lambda$ found via hyperparameter search; 
\textbf{(b)} TVT~\citep{hung2019optimizing} (using the code accompanying the paper);
\textbf{(c)} \notrudder{}: a reward redistribution method \emph{inspired} by RUDDER~\citep{arjona2019rudder};
\textbf{(d)} Meta-$\lambda(s)$~\citep{xu2018meta}: meta-learning a state-dependent function $\lambda(s)$ for $\lambda$-returns; and \textbf{(e)} RGM~\citep{wang2019beyond}: meta-learning a \textit{single} set of weights for generating returns as a linear combination of rewards.

\paragraph{Environment and Parametric Variation.}
KtD is a fixed-horizon episodic task where each episode consists of three phases (Figure~\ref{fig:ktd-layout} top).
In the \emph{Key phase} ($15$ steps in duration) there is no reward and the agent must navigate to the key to collect it. 
In the \emph{Apple phase} ($90$ steps in duration) the agent collects apples; apples disappear once collected. Each apple yields a noisy reward with mean $\mu$ and variance $\sigma^{2}$. 
In the \emph{Door phase} ($15$ steps in duration) the agent starts at the center of a room with a door but can open the door only if it has collected the key  earlier. Opening the door yields a reward of $10$.
Crucially, picking up the key or not has no bearing on the ability to collect apple rewards. The apples are the noisy rewards that \emph{distract the agent from learning that picking up the key early on leads to a door-reward later}. In our experiments, we evaluate methods on $9$ different environments representing  combinations of $3$ levels of apple reward mean and $3$ levels of apple reward variance.

\paragraph{Implementation.} 
The agent observes the top-down view of the current phase rendered in RGB and a binary variable indicating if the agent collected the key or not. 
The policy ($\theta$) and the value functions ($\psi$ and $\phi$) are implemented by separate convolutional neural networks. The \textit{meta-network} ($\eta$) computes the pairwise weight $w_{ij}$ as follows: First, it embeds the observations $s_{i}$ and $s_{j}$ and the time difference $(j - i)$ into separate latent vectors. Then it takes the element-wise product of these three vectors to fuse them into a vector $h_{ij}$. Finally it maps $h_{ij}$ to a scalar output that is bounded to $[0, 1]$ by sigmoid. 
We tuned hyperparameters for each method on the mid-level configuration $\langle \mu=5, \sigma=25 \rangle$ and kept them fixed for the other $8$ configurations.
Each method has a distinct set of parameters (e.g.\ outer-loop learning rates, $\lambda$). We referred to the original papers for the parameter ranges searched over. More details are in the appendix.

\paragraph{Empirical Results.}
\begin{figure}
  \centering
  {\includegraphics[width=\linewidth]{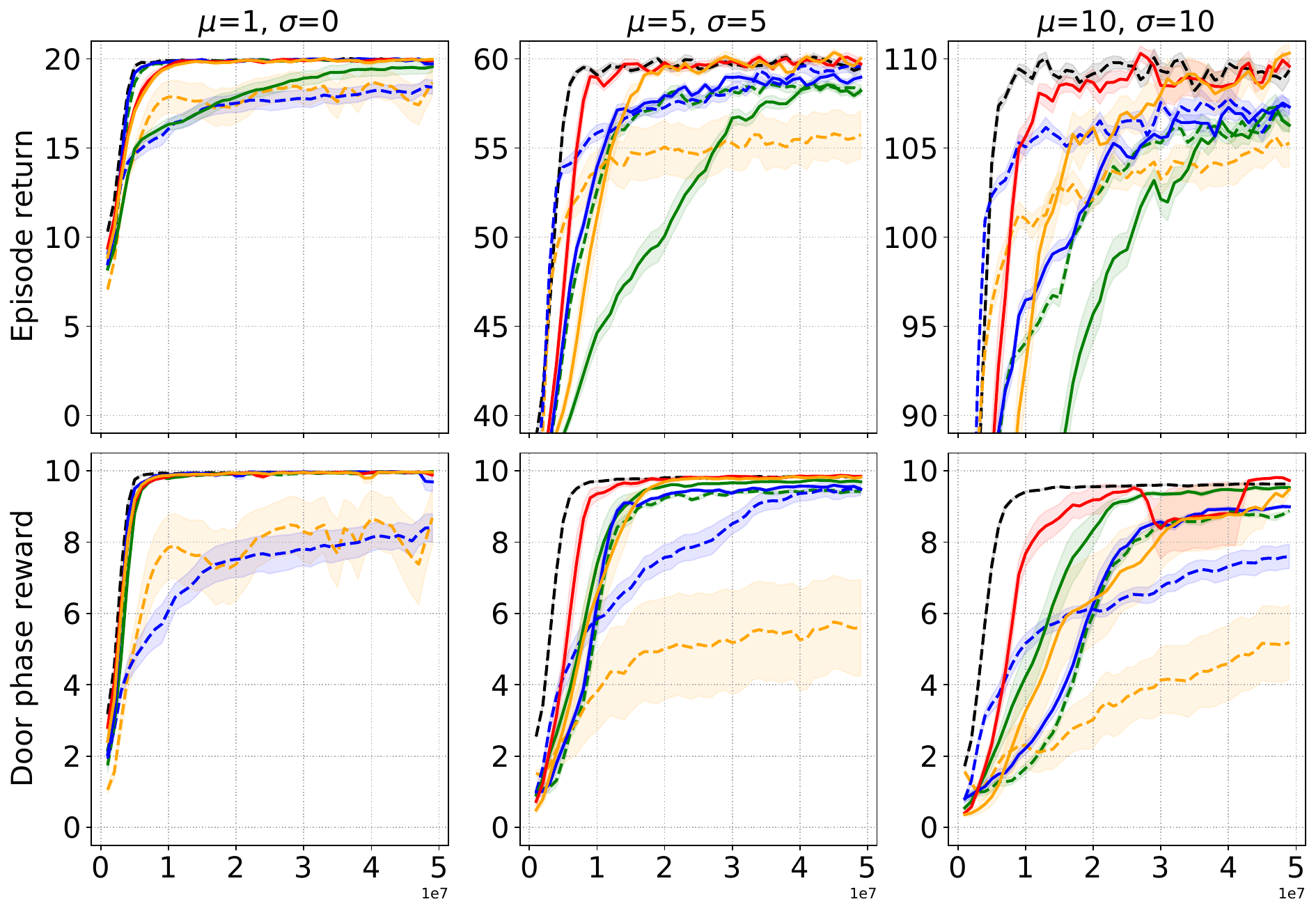}} \quad
  {\includegraphics[width=\linewidth]{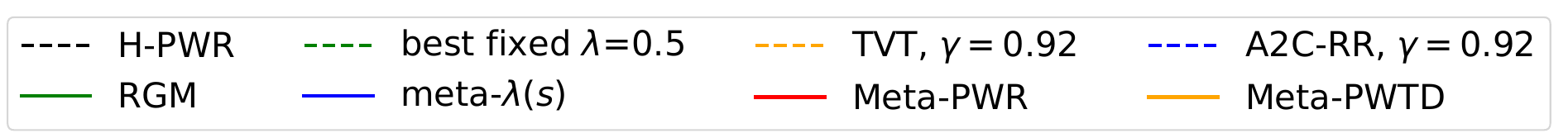}}
  \caption{Learning curves for the KtD domain. Each column corresponds to a different configuration. The x-axis denotes the number of frames. The y-axis denotes the episode return in top row and the door phase reward in bottom row. The solid curves show the average over $10$ independent runs and the shaded area shows the standard errors.}
  \label{fig:ktd-learning-curves}
\end{figure}
Figure~\ref{fig:ktd-learning-curves} presents learning curves for Meta-PWTD, Meta-PWR, and baselines in three KtD configurations (the remaining configurations are in the appendix). 
Learning curves are shown separately for the \emph{total episode return} and the \emph{door phase reward}, the latter a measure of success at the long-term credit assignment.
Not unexpectedly, H-PWR which uses handcrafted pairwise weights performs the best. The gap in performance between H-PWR and the best fixed-$\lambda$ shows that this domain provides a credit assignment challenge that the pairwise-weighted advantage estimate can help with.
The TVT and \notrudder{} methods used a low discount factor and so relied solely on their heuristics for learning to pick up the key, but neither appears to enable fast learning in this domain.
In the door phase, Meta-PWR is generally the fastest learner after H-PWR. Meta-PWTD, though slower, achieves optimal performance. Although RGM performs third best in the door phase, it does not perform well overall, suggesting that the inflexibility of its single set of reward weights (vs.\ pairwise of Meta-PWR) forces a trade off between short and long-term credit assignment.
In summary, Meta-PWR outperforms all the other methods and Meta-PWTD is comparable to the baselines.

Figure~\ref{fig:ktd-layout} presents a visualization of the handcrafted weights for H-PWR (bottom left) and weights learned by Meta-PWR (bottom right). 
In each heatmap, the element on the $i$-th row and the $j$-th column denotes $w_{ij}$, the pairwise weight for computing the contribution of the reward upon transition to the $j$-th state to the return at the $i$-th state in the episode. In the heatmap of the handcrafted weights, the top-right area has non-zero weights because the rewards in the door phase depend on the actions selected in the key phase. The weights in the remaining part of the top rows are zero because those rewards do not depend on the the actions in the key phase. For the same reason, the weights in the middle-right area are zero as well. The weights in the rest of the area resemble the exponentially discounted weights with a discount factor of $0.92$. This steep discounting helps fast learning of collecting apples. 
The learned weights largely resemble the handcrafted weights, which indicate that the metagradient procedure was able to simultaneously learn (1) the important rewards for the key phase are in the door phase, and (2) a quick-discounting set of weights within the apple phase that allows faster learning of collecting apples.


\begin{table*}[tb]
\centering
\begin{tabular}{lccccccc}
\toprule
& Catch & Catch Noise & Catch Scale & Umbr. Length & Umbr. Distract & Cartpole & Discount Chain \\
\hline
A2C & 5975 & 42221 & 56800 & 38050 & 37524 & 76874 & 3554 \\
\notrudder{} & \textbf{5950} & 42295 & 57033 & 38083 & 37433 & 71506 & 3548 \\
RGM & 7849 & 48268 & 54421 & 40397 & 40159 & 119102 & 2444 \\
\hdashline
Meta-PWTD & 6096 & \textbf{41106} & \textbf{48199} & \textbf{37973} & 37226 & 65945 & 1040 \\
Meta-PWR & 5967 & 43076 & 49361 & 38168 & \textbf{36554} & \textbf{61752} & \textbf{161} \\
\bottomrule
\end{tabular}
\caption{Total regret on selected \texttt{bsuite} domains (low is good).}
\label{tab:bsuite}
\end{table*}

\subsection{Experiments on Standard RL Benchmarks}
\label{subsec:bsuite}
Both the DAG and KtD domains are idealized credit assignment problems. However, in domains outside this idealized class, Meta-PWTD and -PWR may learn slower than baseline methods due to the additional complexity they introduce. To evaluate this possibility we compared them to baseline methods on \texttt{bsuite}~\citep{osband2019behaviour} and Atari~\citep{bellemare2013arcade}, both standard RL benchmarks. For these experiments, we did not compare to Meta-$\lambda(s)$ because it performed similarly to the fixed-$\lambda$ baseline in previous experiments as noted in the original paper~\citep{xu2018meta}.

\texttt{bsuite} is a set of unit-tests for RL agents: each domain tests one or more specific RL-challenges, such as exploration, memory, and credit assignment, and each contains several versions varying in difficulty. We selected all domains that are tagged by ``credit assignment'' and at least one other challenge. These domains are not designed solely as idealized credit assignment problems. We ran all methods for $100K$ episodes in each domain except \emph{Cartpole}, which we ran for $50K$ episodes. Each run was repeated $3$ times with different random seeds. Table~\ref{tab:bsuite} shows the total regret.
Overall Meta-PWTD or -PWR achieved the lowest total regret in all domains except for \emph{Catch}. 
It shows that Meta-PWTD and -PWR perform better than or comparably to the baseline methods even in domains without the idealized umbrella-like TCA structure.

To test scalability to high-dimensional environments, we conducted experiments on Atari.
Atari games often have long episodes of more than $1000$ steps thus episode truncation is required. 
However, the returns in RGM and Meta-PWR are not in a recursive additive form thus the common way of correcting truncated trajectories by bootstrapping from the value function is not applicable. 
TVT also requires full episodes for value transportation. 
Therefore, we excluded RGM, TVT, and Meta-PWR and only ran Meta-PWTD, \notrudder{} and A2C.
For each method we conducted hyperparameter search on a subset of $6$ games and ran each method on $49$ games with the fixed set of hyperparameters; see appendix for details.
An important hyperparameter for the A2C baseline is $\lambda$, which was set to $0.95$. 

\begin{figure}[t]
  \centering
  {\includegraphics[width=\linewidth]{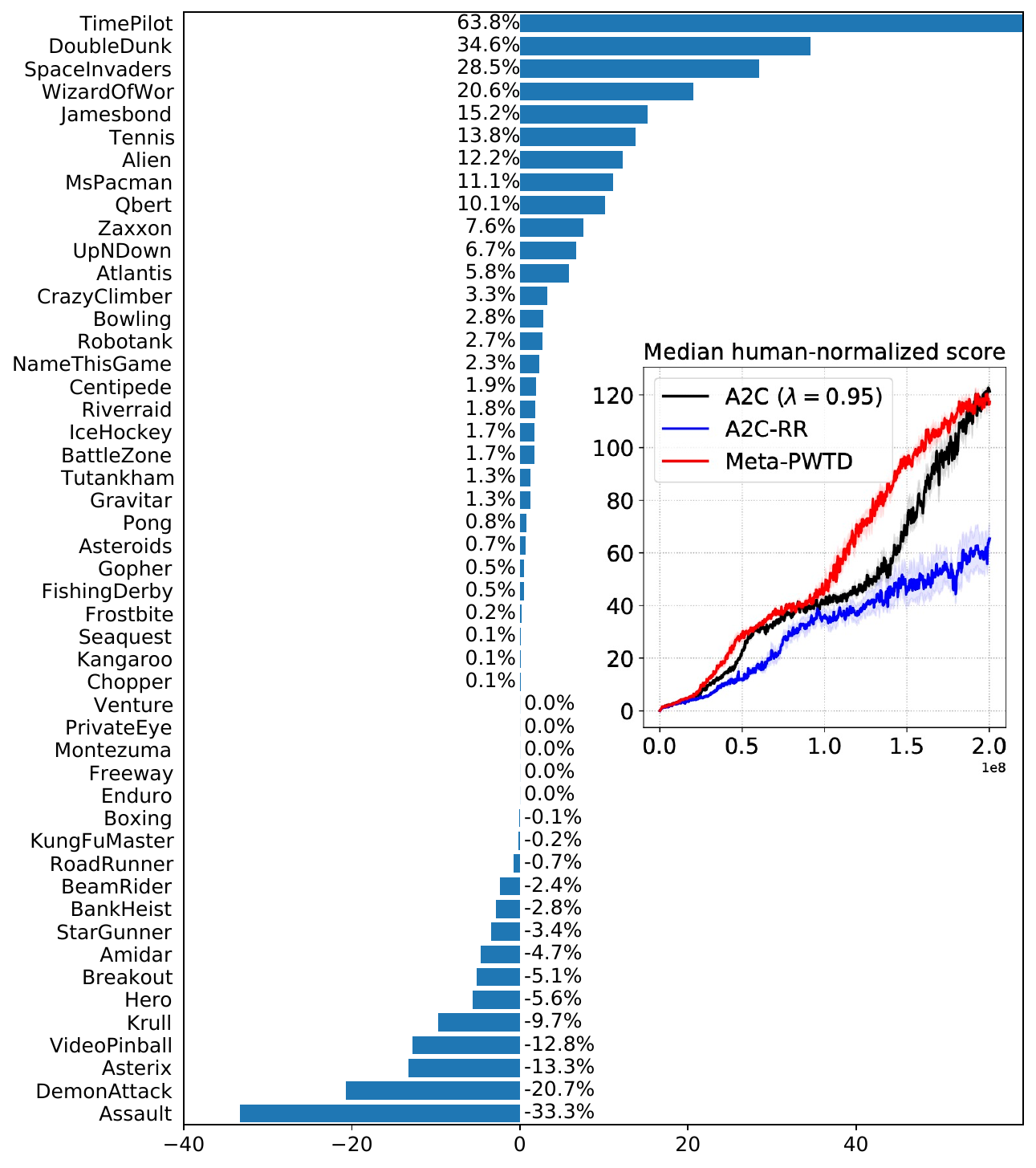}}
  \caption{Relative performance of Meta-PWTD over A2C ($\lambda=0.95$). All scores are averaged over $5$ independent runs with different random seeds. Inset: Learning curves of median human normalized score of all $49$ Atari games. Shaded area shows the standard error over $5$ runs.}
  \label{fig:atari-bar}
\end{figure}

Figure~\ref{fig:atari-bar} (inset) shows the median human-normalized score during training. Meta-PWTD performed slightly better than A2C over the entire period, and both performed better than \notrudder{}
Figure~\ref{fig:atari-bar} shows the relative performance of Meta-PWTD over A2C. Meta-PWTD outperforms A2C in $30$ games, underperforms in $14$, and ties in $5$.
These results show that Meta-PWTD can scale to high-dimensional environments like Atari. 
We conjecture that Meta-PWTD provides a benefit in games with embedded umbrella problems but this is hard to verify directly. 

\section{Conclusion}\label{sec:conclusion}
We presented two new advantage estimators with pairwise weight functions as parameters to be used in policy gradient algorithms, and a metagradient algorithm for learning the pairwise weight functions. 
Simple analysis and empirical work confirmed that the additional flexibility in our advantage estimators can be useful in domains with delayed consequences of actions, e.g., in umbrella-like problems. 
Empirical work also confirmed that the metagradient algorithm can learn the pairwise weights fast enough to be useful for policy learning, even in large-scale environments like Atari. 


\clearpage

\section*{Acknowledgements}
This work was supported by DARPA's L2M program as well as a grant from the Open Philanthropy Project to the Center for Human Compatible AI. 
Any opinions, findings, conclusions, or recommendations expressed here are those of the authors and do not necessarily reflect the views of the sponsors.

\bibliography{aaai22}

\clearpage
\appendix

\section{Metagradient Algorithm}
In this section we give further details on the metagradient algorithm used to learn the pairwise weights for both Meta-PWTD and Meta-PWR as well as the metagradient baselines. We use the algorithm from \citep{xu2018meta}. A generic metagradient algorithm is presented in Algorithm \ref{alg:metagrad} where we have extended the syntax from the paper to include the trajectories and meta-parameters in the inputs of the objective functions to make the dependencies more explicit. For clarity, the algorithm omits the value function update steps, which differ between the specific algorithms as explained in the paper but do not change the computation of metagradients. In our practical implementation we do not use the explicit gradient expressions given here. Instead we define a computational graph that includes the inner update and the outer update and compute metagradients via automatic differentiation.

The algorithm updates both the policy parameters $\theta$ and the meta-parameters $\eta$ iteratively. In the $k$-th iteration, it updates the policy parameters $\theta_k$ using the gradient of the inner objective $J^{\text{inner}}(\theta_k, \tau_k; \eta_k)$, which is parametrized by the meta-parameters $\eta_k$. The updated parameters are used for sampling another batch of trajectories $\tau_{k+1}$. The meta-objective $J^{\text{outer}}(\theta_{k+1}, \tau_{k+1})$ is computed as the standard policy gradient loss on the new trajectory. The updated parameters are a differentiable function of the meta-parameters, which enables the computation of metagradients as given in Equation $15$. The gradient of the updated parameters with respect to the meta-parameters can be written as follows
\begin{align*}
  \nabla_\eta \theta_{k+1} =& \nabla_{\eta} \big( \theta_{k} + \alpha^{\theta} \nabla_{\theta} J^{\text{inner}}(\theta_k, \tau_k; \eta_k) \big) \\
  \approx & \alpha^{\theta} \nabla_{\eta} \nabla_{\theta} J^{\text{inner}}(\theta_k, \tau_k; \eta_k) \\
  =& \alpha^{\theta} \big( \nabla_{\eta} \nabla_{\theta}J_{\eta}(\theta_k, \tau_k) + \beta^{\mathcal{H}} \nabla_{\eta} \nabla_{\theta} \mathcal{H}(\pi_{\theta_k}) \big) \\
  =& \alpha^{\theta} \nabla_{\eta} \nabla_{\theta}J_{\eta}(\theta_k, \tau_k)
\end{align*}
The second equation is a greedy approximation of the metagradients by dropping the dependency of the parameter $\theta_k$ on the previous updates (cf.~\citep{xu2018meta}). The fourth equation is because the entropy regularization term does not depend on the meta-parameters $\eta$. 
The approximated metagradients can be estimated from sampled trajectories as
\begin{align*}
  \nabla_\eta \theta_{k+1} \approx \alpha^{\theta} \sum_{\tau_{k} \sim \pi_{\theta_{k}}} \Big[\sum_{t=0}^{T-1} \nabla_\eta \hat{\Psi}_{\eta}(S_t, A_t) \nabla_{\theta}\log \pi_{\theta}(A_{t}|S_{t})\Big].
\end{align*}
Dropping the dependency of the parameters on the previous updates introduces bias to the gradient. We leave the investigation of using an unbiased gradient estimator for future work.

\begin{figure}
  \centering
  \begin{algorithm}[H]
      \centering
      \label{alg:algorithm}
      \caption{A generic metagradient algorithm.}
      \begin{algorithmic}
          \STATE \textbf{Input} Initial parameters $\theta_0$ and $\eta_0$.
          \STATE Sample a batch of trajectories $\tau_0$ with $\theta_0$;
          \REPEAT
              \STATE Set $\theta_{k+1} = \theta_{k} + \alpha^{\theta} \nabla_{\theta} J^{\text{inner}}(\theta_k, \tau_k; \eta_k)$;
              \STATE Sample a batch of trajectories $\tau_{k+1}$ with $\theta_{k+1}$;
              \STATE Set $\eta_{k+1} = \eta_{k} + \alpha^{\eta} \nabla_{\eta} J^{\text{outer}}(\theta_{k+1}, \tau_{k+1})$;
              \STATE $k \leftarrow k + 1$;
          \UNTIL{done}
      \end{algorithmic}
      \label{alg:metagrad}
  \end{algorithm}
\end{figure}

\section{DAG Experiments}
Three algorithms were compared in a tabular domain. In this appendix, the hyperparameter configurations of the three algorithms are provided in Section~\ref{subsec:dag-hyperparameters} and more detailed results are reported in Section~\ref{subsec:dag-results}.

\subsection{Hyperparameters}
\label{subsec:dag-hyperparameters}
Fixed-$\lambda$ uses the Adam optimizer~\citep{kingma2014adam} with learning rate $0.01$, $\beta_1 = 0$, $\beta_2 = 0.999$, and $\epsilon=10^{-8}$. 
The learning rate was selected from a coarse-grained search in $\{0.1, 0.01, 0.001\}$. The other Adam parameters were chosen by hand without search. 
Updates are computed on batches consisting of $8$ full episodes.
We found $\lambda=1$ the best, better than any smaller values.
We used a discount factor $\gamma=1$ and an entropy regularization coefficient $0.001$.
The inner loop of Meta-PWTD, Meta-PWR, and H-PWR share hyperparameters with the fixed-$\lambda$ baseline, except that $\lambda$ is not used.
For Meta-PWTD and Meta-PWR, the outer-loop optimizer is Adam with the same hyperparameters as the inner-loop optimizer.
Outer-loop gradient is clipped to $0.5$ by global norm.
The weight matrix $\eta$ is initialized from uniform distribution in range $[-0.01, 0.01]$.

\subsection{Additional Empirical Results}
\label{subsec:dag-results}
\begin{figure*}
    \centering
    \includegraphics[width=0.8\textwidth]{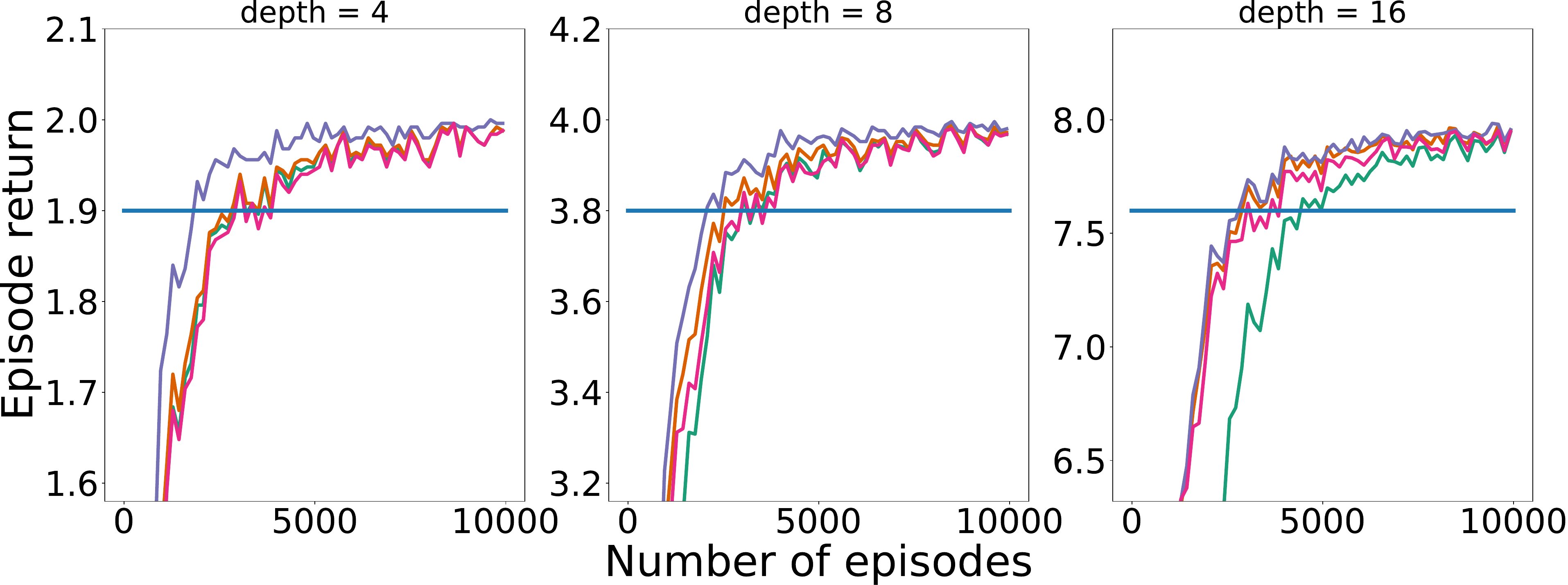}
    \includegraphics[width=0.8\textwidth]{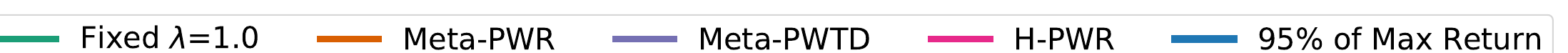}
    \caption{Learning curves in the DAG environments. Each curve is the mean of $5$ seeds. The line for $95\%$ of max return is added to help contextualize these learning curves with Figure 2 in the main paper.}
    \label{fig:dag-curves}
\end{figure*}

Learning curves in the DAG environment are presented in Figure~\ref{fig:dag-curves}. All methods converged to the optimal performance by the end of $10000$ episodes of training. 
Handcrafted and meta-learned weights in the depth-$4$, depth-$8$, and depth-$16$ DAG environment variants are presented in Figures~\ref{fig:supp-depth-4-dag-weights}, \ref{fig:supp-depth-8-dag-weights}, and \ref{fig:supp-depth-16-dag-weights}.

\subsection{Learning TD-error Weights with Different Value Functions}
\label{subsection:masked-value-td-weights}
\begin{figure*}
  \centering
  \subfloat[H-PWR]{\includegraphics[width=0.25\textwidth]{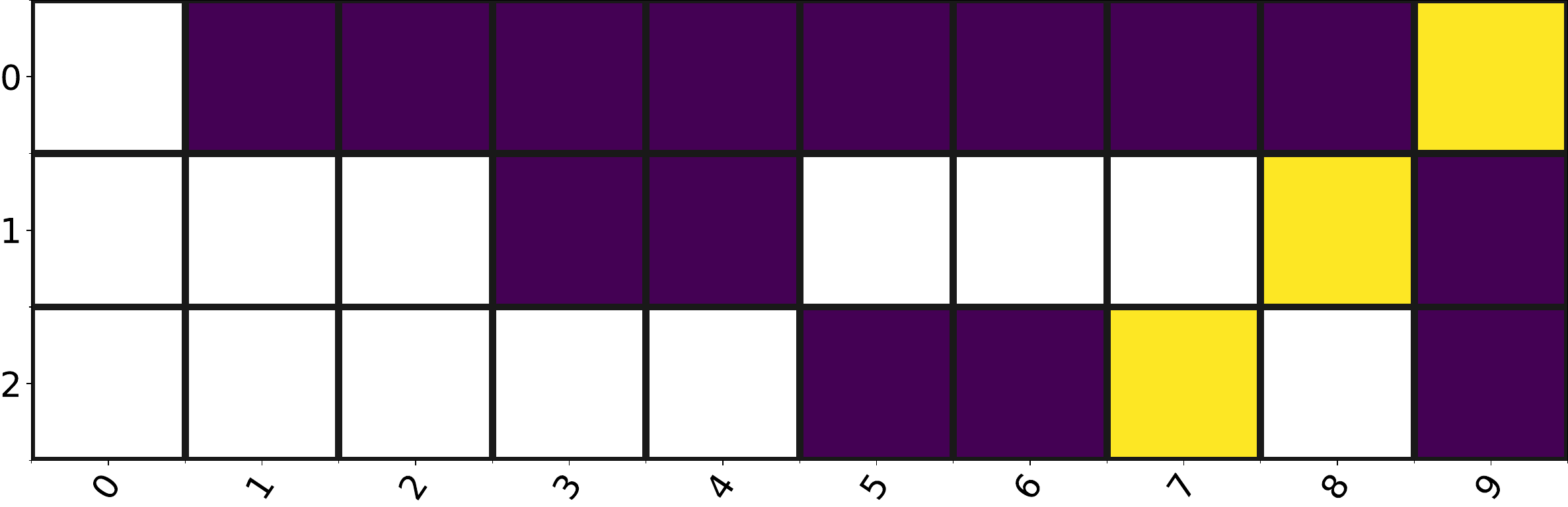}}
  \subfloat[Meta-PWR]{\includegraphics[width=0.25\textwidth]{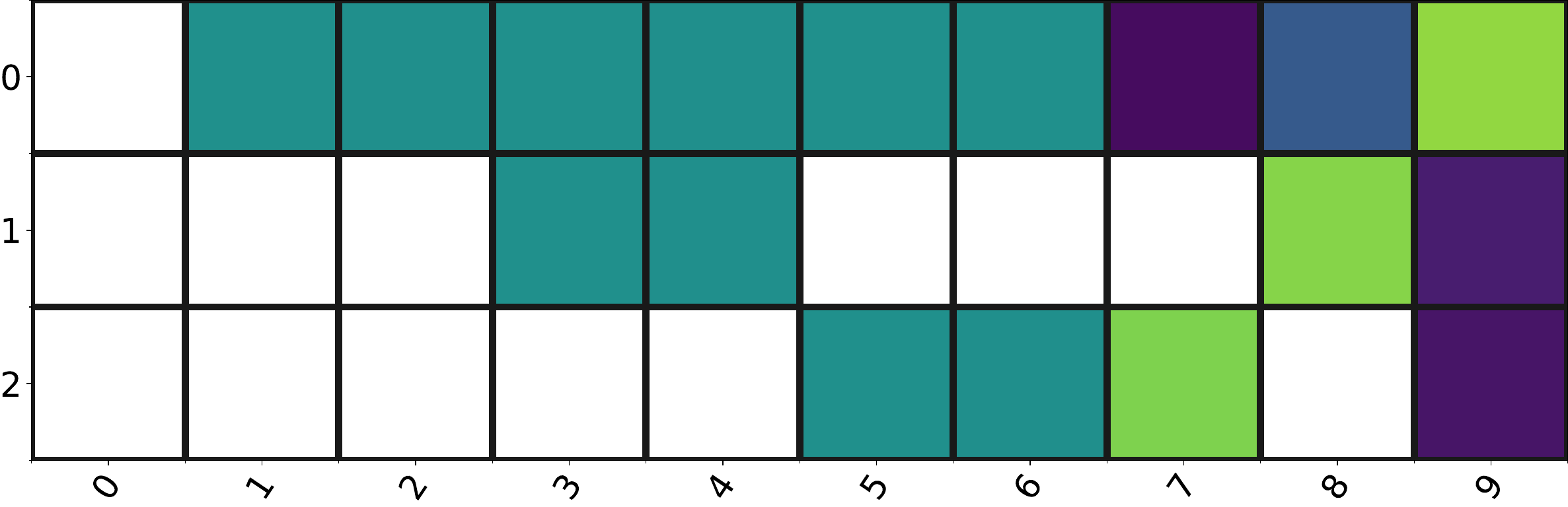}}
  \subfloat[Meta-PWTD]{\includegraphics[width=0.25\textwidth]{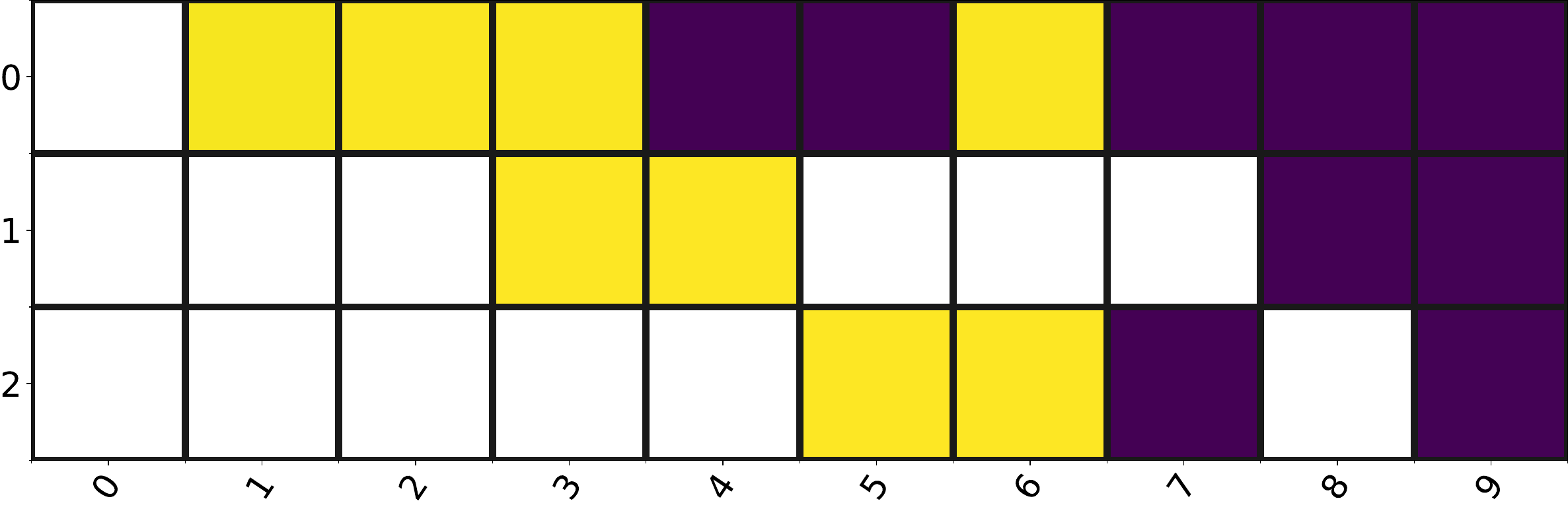}} \quad
  \subfloat{\includegraphics[width=0.5\textwidth, trim=0 0 10 0]{figures/1d_reset/supplementary_weight_matrices/supplementary_horizontal_colorbar.pdf}}
  \caption{Handcrafted, meta-learned reward weights, and meta-learned TD-error weights in the depth-$4$ DAG environment in \textbf{a}, \textbf{b}, and \textbf{c} respectively. White is unreachable. Y-axis has been cropped to only include weights that affect inner loop learning.}
  \label{fig:supp-depth-4-dag-weights}
\end{figure*}

In the main paper we show that the weights learned by Meta-PWR in the $\theta$-reset experiment converge to a fixed weight matrix, which resemble the handcrafted weights that we believe are useful for variance reduction (see Figure~\ref{fig:supp-depth-4-dag-weights}, Figure~\ref{fig:supp-depth-8-dag-weights}, and Figure~\ref{fig:supp-depth-16-dag-weights} for visualizations of the weights). 
There exist sets of fixed weights that are beneficial throughout the learning process because the reward associated with each transition is stationary. 
In contrast, the TD-error for each transition is non-stationary and evolves as the policy and the value function change. Accordingly, the weights learned by Meta-PWTD are also non-stationary. 
To visualize and understand the weights learned by Meta-PWTD, we conducted a set of experiments where we set the value function by hand and held it fixed during learning so that the weights could converge to a fixed point. Specifically, we use the optimal value function but set the values to zero for all states up to certain depth and hold it fixed during learning. 

In Figure~\ref{fig:dag_masked_value}, three cases of value masking are shown in the depth-$8$ DAG environment where the optimal value function has been masked to depth $0$, $4$, and $8$. 
In the mask depth $0$ case, the weights for the TD-errors have mostly changed in the parts of the weight matrix before column $30$. Column $30$ corresponds to the last state of the middle-layer of the DAG that splits the states where the agent can act from the states where the agent receives rewards. 
When values are masked up to depth $4$, the TD-error weighting shifts. State $14$ is the last state at depth $4$, so all weights before that are unchanged from the initialization. Note that at mask depths $0$ and $4$, no weights are placed on the states in the weight matrix after column $30$.
Finally, at depth $8$, all of the value function has been masked out and the TD-error weighting has converged to visually similar weights as the weights learned by Meta-PWR, which is expected as the TD-errors computed with the fully masked value function consist only of the rewards.
The learned weights in Figure~\ref{fig:dag_masked_value} show that Meta-PWTD learns different kinds of weightings depending on the value function.

\section{Key-to-Door Experiments}
\subsection{Environment Description}
Key-to-Door (KtD) is a fixed-horizon episodic task where each episode consists of three 2D gridworld phases. 

In the \emph{Key phase} (duration $15$ steps), there is no reward and the agent must navigate in a $5 \times 5$ map to collect a key. The key disappears once collected. 
The initial locations of the agent and the key are randomly sampled in each episode.

In the \emph{Apple phase} (duration $90$ steps), the agent collects apples in a $5 \times 9$ map by walking over them; apples disappear once collected. 
Each apple yields a noisy reward with mean $\mu$ and variance $\sigma^{2}$. Specifically, each apple yields a reward of $r = \mu \tau$ with probability $\frac{1}{\tau}$ or a reward of $0$ with probability $1 - \frac{1}{\tau}$ where $\tau = \frac{\sigma^{2}}{\mu^{2}} + 1$. This sampling procedure is consistent with the original TVT paper~\citep{hung2019optimizing}.
The number of apples is uniformly sampled from $[1, 20]$ and their locations are randomly sampled.

In the \emph{Door phase} (duration $15$ steps), the agent starts at the center of a $3 \times 3$ room with a door. The agent can open the door only if it has collected the key in the earlier Key phase. The door disappears after being opened. Successfully opening the door yields a reward of $10$. 

The agent's observation is a tuple, $(map, has\_key)$. $map$ is the top-down view of the current phase and is rendered in an RGB representation. $has\_key$ is a binary channel which is $1$ if the agent has already collected the key and $0$ otherwise. The agent has $4$ actions which correspond to moving \emph{up}, \emph{down}, \emph{left}, and \emph{right}. The primary difference between our KtD environment and the original is that our environment is fully observable; the original is partially observable. This difference is reflected in two modifications: the agent observes the top-down view of the map rather than the first-person view, and the agent observes whether it has collected the key.

We also conducted experiments in a stochastic variant of KtD to test the robustness of Meta-PWR and Meta-PWTD to stochastic transition dynamics. In the stochastic variant, the action being executed is replaced by a random action with probability $0.1$ for each time step.

\subsection{Implementation Details}
\label{subsec:ktd-architecture}
All methods use A2C~\citep{mnih2016asynchronous} as the policy optimization algorithm. $16$ actors are used to generate data. The rollout length is equal to the episode length, $120$ in this case.
For each method described below, we conducted a hyperparameter search in the $\mu=5$ and $\sigma=5$ configuration and selected the best-performing hyperparameters. Then the hyperparameters were fixed for all the other $8$ environment configurations.
Each candidate hyperparameter combination was run with three different random seeds for $50$ million frames. The best hyperparameter combination was determined to be the one that first achieved $57$ in episode return, i.e., $95\%$ of the maximum possible.
The following hyperparameter settings are shared across all methods unless otherwise noted:
learning rate $2 * 10^{-4}$, and Adam $\beta_1=0$, Adam $\beta_2=0.999$, Adam $\epsilon=10^{-8}$, discount factor $\gamma=0.998$, and entropy regularization coefficient $0.05$.
The advantage estimates are standardized in a batch of trajectories before computing the policy gradient loss~\citep{baselines} unless otherwise noted.

\paragraph{A Standard Perception module.}
A standard perception module is used by all method to process the observation $s$ to a latent vector $h$.
The observation $s$ is a tuple, $(map, has\_key)$. $map$ is the top-down view of the current phase that has shape $(7, 11, 3)$, where the last dimension is the RGB channels. $has\_key$ is a binary channel which is $1$ if the agent has already collected the key and $0$ otherwise.
$map$ is processed by two convolutional layers with $16$ and $32$ filters respectively. Both convolutional layers use $3\times3$ kernels and are followed by ReLU activation.
The output of the last convolutional layer is then flattened and processed by Dense(512) - ReLU.
The binary input $has\_key$ is concatenated with the ReLU layer output. 
Finally, the concatenated vector is further processed by a MLP: Dense(512) - ReLU - Dense(256) - ReLU.
We denote the final output of the perception module as $h$.

\paragraph{Fixed-$\lambda$}
The fixed-$\lambda$ baseline implements the standard A2C algorithm. The policy and value function are implemented by two separate neural networks consisting of a perception module and an output layer without any parameter sharing. The policy network maps $h$ to the policy logits via a single dense layer. The value network maps $h$ to a single scalar via a single dense layer.
We label this baseline fixed-$\lambda$ to underline the importance of the eligibility trace parameter $\lambda$.
We searched for all combinations of the following hyperparameter sets: $\lambda$ in $\{1.0, 0.99, 0.98, 0.95, 0.9, 0.8, 0.5, 0\}$, and learning rate in $\{10^{-3}, 2*10^{-4}, 5*10^{-5}, 10^{-5}\}$.
The best performing set of hyperparameters is $\lambda = 0.5$ and learning rate $2*10^{-4}$.

\paragraph{Meta-PWTD}
The policy ($\theta$) and value function for the original return ($\psi$) have the same network architecture as in the fixed-$\lambda$ baseline. 
The meta-network ($\eta$) computes the weights as follows. 
For each episode, the inputs to the meta-network is a sequence $(s_{0}, \delta_{1}, s_{1}, \dots, \delta_{T}, s_{T})$. Note that the TD-error $\delta_{i}$ is part of the inputs.
The meta-network first maps each $s_{t} (0 \le t \le T)$ into a latent vector $h_{t}$ with a standard perception module. The meta-network ($\eta$) shares the perception module with the value function for the original return ($\psi$). No gradient is back-propagated from the meta-network to the shared perception module.
A dense layer with $256$ hidden units maps $h_{i} (0 \le i < T)$ into $h^{row}_{i}$; a separate dense layer with $256$ units maps $h_{j} \oplus \delta_{j} (0 < j \le T)$ into $h^{col}_{j}$ where $\oplus$ denotes concatenation. The TD-error $\delta_{j}$ is clipped to $[-1, 1]$ before concatenation. Both dense layers are followed by ReLU activation. Another dense layer with $256$ units maps the time interval $(j - i) (0 \le i < j \le T)$ to a latent vector $td_{ij}$. $h^{row}_{i}$, $h^{col}_{j}$, and $td_{ij}$ are element-wise multiplied to fuse the three latent vectors into one vector $h_{ij}$: $h_{ij} = (h^{row}_{i} + 1) * (h^{col}_{j} + 1) * (td_{ij} + 1)$. Note that every vector is shifted by a constant $1$ before the multiplication to mitigate gradient vanishing at the beginning of training~\citep{perez2018film}.
The latent vectors $h_{ij} (0 \le i < j \le T)$ are normalized by
\[
    h'_{ijd} = \gamma_{d} \frac{(h_{ijd} - \mu_{d})}{\sigma_{d}} + \beta_{d}\quad(1 \le d \le 256),
\]
where 
\[
    \mu_{d} = \frac{2}{T * (T + 1)} \sum_{0 \le i < j \le T} h_{ijd}
\]
and
\[
    \sigma_{d} = \sqrt{\frac{2}{T * (T + 1)} \sum_{0 \le i < j \le T} (h_{ijd} - \mu_{d})^2}
\]
are the empirical mean and standard deviation of $h_{ij}$ respectively. $\gamma_{d}$ and $\beta_{d}$ are trainable parameters. ReLU activation is applied to $h'_{ij}$.
Finally, the output layer maps each $h'_{ij}$ into $w_{ij}$. The initial weights for the output layer is scaled by a factor $0.01$ so that the initial outputs are closer to uniform. We applied sigmoid activation on the outputs to bound the weights to $(0, 1)$.
As for hyperparameters, the entropy regularization coefficient is set to $0.05$. For the inner loop, we used the Adam optimizer with learning rate $2 * 10^{-4}$, $\beta_{1} = 0$, $\beta_{2} = 0.999$, and $\epsilon = 10^{-8}$. For the outer loop, we used the Adam optimizer with learning rate $2 * 10^{-5}$, $\beta_{1} = 0$, $\beta_{2} = 0.999$, and $\epsilon = 10^{-8}$.

\paragraph{Meta-PWR}
Meta-PWR uses the same neural network architecture as Meta-PWTD. The only difference is that Meta-PWR does \emph{not} take the reward $r_{i}$ or the TD-error $\delta_{i}$ as inputs. 
In addition, Meta-PWR employs a value function for the weighted sum of rewards ($\phi$) which has an identical architecture as the value function for the original return ($\psi$). 
The hyperparameters for Meta-PWR are also the same as those for Meta-PWTD.

\paragraph{H-PWR}
The policy and value function for the weighted sum of rewards have the same network architecture as in the fixed-$\lambda$ baseline. 
We handcraft pairwise weights for the KtD domain to take advantage of the known credit assignment structure that can be described as follows: The policy learning in the key phase depends only on the reward in the door phase, the policy learning in the apple phase does not depend on the other phases, and while the picking up the key or not impacts the reward in the door phase, the reward in the door phase is still instantaneous.
The weights are set so that in the key phase, the rewards in the apple phase receive a zero weight and the rewards in the door phase are discounted starting from the first timestep of the door phase.
Weights that compute discounting equivalent to $\gamma=0.92$ are applied in the apple phase and door phase.
An illustration of the learned weights is presented in Figure 3 in the main paper.
H-PWR uses the same hyperparameters as fixed-$\lambda$ except $\gamma$, which does not apply and $\lambda$, which is set to $1.0$. No hyperparameter search is conducted specifically for H-PWR.

\paragraph{Meta-$\lambda(s)$}
The policy ($\theta$) and value function for the original return ($\psi$) have the same network architecture as in the fixed-$\lambda$ baseline. 
The value function for the weighted sum of rewards ($\phi$) has identical architecture as the value function for the original return ($\psi$). 
The meta-network ($\eta$) maps a state $s_{t}$ to a scalar $\lambda(s_{t}) \in (0, 1)$. The meta-network first maps the observation $s_{t}$ to a latent vector $h_{t}$ with the standard perception module and then maps $h_{t}$ to a single scalar $\lambda(s_{t})$ via a single dense layer. Sigmoid is applied to the output of the meta-network to bound it to $(0, 1)$.
We searched for the outer-loop learning rate in $\{10^{-4}, 2*10^{-5}, 5*10^{-6}, 10^{-6}\}$.
The best performing outer-loop learning rate is $10^{-6}$.
The outer-loop $\lambda$ is set to $1.0$ without search.

\paragraph{RGM}
The policy ($\theta$) and value function for the original return ($\psi$) have the same network architecture as in the fixed-$\lambda$ baseline. 
The value function for the weighted sum of rewards ($\phi$) has identical architecture as the value function for the original return ($\psi$). 
For an episode $\tau = (s_{0}, a_{0}, r_{1}, s_{1}, \dots, s_{T})$, the meta-network first maps each $s_{t}\enspace(0 \le t \le T)$ to $h^{0}_{t}$ by a shared standard perception module. Then it concatenates $h^{0}_{t}$ with $r_{t}$ and the one-hot representation of $a_{t}$. Four Transformer blocks~\citep{vaswani2017attention} are applied on the concatenated features, each block with four attention heads. We denote the output of the final Transformer block as $h^{4}_{t}\enspace(0 \le t \le T)$. Finally, a shared linear layer maps each $h^{4}_{t}$ to $\beta_{t}$, the weight on the reward $r_{t}$.
We searched for the outer-loop learning rate in $\{10^{-4}, 10^{-5}, 10^{-6}, 10^{-7}\}$, as suggested by the original paper~\citep{wang2019beyond}.
The best performing outer-loop learning rate is $10^{-6}$.
The outer-loop $\lambda$ is set to $1.0$ without search.

\paragraph{TVT}
We adopted the agent architecture from the open-source implementation accompanying the original TVT paper (\href{https://github.com/deepmind/deepmind-research/tree/master/tvt}{https://github.com/deepmind/deepmind-research/tree/master/tvt}). The only difference is that we replaced the convolutional neural network torso in the original code with the standard perception module.
We searched for all combinations of the following hyperparameter sets: the read strength threshold for triggering the splice event in $\{1, 2\}$ and the learning rate in $\{10^{-3}, 2*10^{-4}, 5*10^{-5}, 10^{-5}\}$.
The best performing set of hyperparameters are $1$ for the read strength threshold and $2*10^{-4}$ for the learning rate.
We did not use advantage standardization for TVT because we found that it hurt the performance in the KtD domain.
We used the Adam optimizer parameters $\beta_{1} = 0$, $\beta_{2} = 0.95$, and $\epsilon = 10^{-6}$, as the open-source implementation suggested.
We also set $\lambda=0.92$ and $\gamma=0.92$ following the original implementation.

\paragraph{\notrudder{}}
We pick the main ideas from RUDDER~\cite{arjona2019rudder} and implement them in an
algorithm we call \notrudder{}.
RUDDER uses contribution analysis to redistribute rewards in a RL
episode. The high-level idea is that since the environment rewards may be
delayed from the transitions that resulted in them, contribution analysis may
be used to compute how much of the total return is explained by any
particular transition. In effect, this drives the expected future return to
zero because any reward that can be expected at any given timestep will be
included in the redistributed reward, eliminating the delay and leading to
faster learning of the RL agent. The method is based on learning a
LSTM-network, which predicts the total episodic return at every timestep. The
redistributed reward is computed as the difference of return predictions on
consecutive timesteps.

Compared to the full RUDDER algorithm, \notrudder{} incorporates a few changes to isolate some of the core ideas of the reward redistribution module and make the algorithm more directly comparable to the other A2C-based algorithms discussed in this paper.
We use the LSTM cell-architecture proposed in the RUDDER paper and train it
from samples stored in a replay buffer. In the KtD-experiments, we do not use
the ``quality'' weighted advantage estimate proposed in the paper due to the
random noise in the episodic return, which we deem too high variance for
reliable estimation of the redistribution quality. Instead we mix the
original and redistribution-based advantages at a fixed ratio.
We recognize that omitting some of the
features of the full RUDDER algorithm may adversely impact the reward
redistribution and therefore the agent learning performance. Nevertheless, we
believe the reward redistribution idea is an interesting take on a similar
idea as the pairwise weighting studied in this paper and therefore provide
our implementation -- \notrudder{} -- of that idea as a baseline.

The regular frames and delta frames (as described in the paper) are processed by the standard perception module.
The preception module outputs and the one-hot encoded action are concatenated and processed by Dense(512) - ReLU.
The output of the ReLU layer is the input for the reward redistribution model.
As suggested in the paper, the reward redistribution model is a LSTM without a forget gate and output gate.
The cell input only receives forward connections and the gates only receive recurrent connections.
All of the layers in the LSTM have $64$ units.
We chose not to use the prioritized replay buffer described in the paper due to the high variance of the returns in the KtD environment.
For the same reason we did not use the quality measure, which is also described in the paper, for mixing the RUDDER advantage and regular advantage.
Instead, we used a fixed mixing coefficient, which we searched for.
The advantage is standardized after the mixing.
We implemented an auxiliary task described in the paper, where the total return prediction loss is applied at every step of the episode.
The reward redistribution model is trained for $10$ randomly sampled batches of size $8$ from a circular buffer holding the past $128$ trajectories between each policy update.
We set the number of updates to $10$ via an informal hyperparameter search in the $\mu=5$, $\sigma=5$ KtD setting, where we found that training $10$ times between each update performs better than $5$ but further increasing it did not yield further large improvements.
The reward redistribution model is trained with Adam with learning rate $10^{-4}$, $\beta_1=0.9$, and $\beta_2=0.999$.
We applied a L2 weight regularizer with coefficient $10^{-7}$.
We searched for all combinations of the following hyperparameter sets: $\gamma$ in $(0.92, 1.0)$, $\lambda$ in $(1.0, 0.95, 0.5)$, auxiliary task coefficient in $(0.0, 0.5)$, and advantage mixing coefficient in $(0.5, 1.0)$.
The best performing set of hyperparameters is $\gamma = 0.92$, $\lambda = 0.5$, auxiliary task coefficient $0.0$, and advantage mixing coefficient $0.5$.

\subsection{Additional Empirical Results}
We ran all of the methods described above in $9$ variants of the KtD environment. Figure~\ref{fig:ktd-episode-all}, Figure~\ref{fig:ktd-door-all}, and Figure~\ref{fig:ktd-apple-all} shows the episode return, the total reward in the door phase, and the total reward in the apple phase respectively.

Noticing that the KtD domain has deterministic dynamics, we also conducted experiments on a stochastic KtD domain where the action being executed is replaced by a random action with probability $0.1$ for each time step. The corresponding results are presented in Figure~\ref{fig:ktd-sto-episode-all}, Figure~\ref{fig:ktd-sto-door-all}, and Figure~\ref{fig:ktd-sto-apple-all}. In general, Meta-PWTD and Meta-PWR still perform better than the baseline methods regardless of the stochasticity in the transition dynamics.

\section{\texttt{bsuite} Experiments}
\subsection{Environment Description}
We selected $7$ tasks which were associated with the ``credit assignment'' tag from \texttt{bsuite}. They present a variety of credit assignment structures, including the umbrella problem in \S $3$ in the main text. Additionally, all domains except \textit{Discount Chain} have multiple tags which create additional challenges than temporal credit assignment. 
We ran all different variants of every task, each with $3$ different random seeds. Unlike the standard data regime of \texttt{bsuite}, we ran each task for $100K$ episodes for all methods to calculate the total regret score, except \textit{Cartpole}, which we ran for $50K$ episodes. We refer the readers to the original \texttt{bsuite} paper~\citep{osband2019behaviour} and the accompanying github repository (\href{https://github.com/deepmind/bsuite}{https://github.com/deepmind/bsuite}) for further details.

\subsection{Implementation Details}
Most methods use a similar neural network architecture as described in~\S\ref{subsec:ktd-architecture}. There are two common differences. First, we used $1$ single actor instead of $16$ parallel actors for generating data. Second, the standard perception module is replaced by a $2$-layer MLP with $64$ hidden units and ReLU activation each layer, because the inputs are vectors instead of images in \texttt{bsuite}. Further architecture differences and hyperparameters are described below.

\paragraph{Actor-critic baseline.} The entropy regularization weight is set to $0.05$. we used the Adam optimizer with learning rate $3 * 10^{-4}$, $\beta_{1} = 0$, $\beta_{2} = 0.999$, and $\epsilon = 10^{-8}$.

\paragraph{Meta-PWTD and Meta-PWR}
Besides the perception module, there are two more differences with~\S\ref{subsec:ktd-architecture}. First, the meta-network ($\eta$) and the value function for the original return ($\psi$) use separate perception module instead of sharing. Second, all the hidden layers after the perception module use $64$ hidden units instead of $256$.
The entropy regularization weight is set to $0.05$. For the inner loop, we used the Adam optimizer with learning rate $3 * 10^{-4}$, $\beta_{1} = 0$, $\beta_{2} = 0.999$, and $\epsilon = 10^{-8}$. For the outer loop, we used the Adam optimizer with learning rate $3 * 10^{-5}$, $\beta_{1} = 0$, $\beta_{2} = 0.999$, and $\epsilon = 10^{-8}$. Outer-loop gradient is clipped to $0.01$ by global norm.

\paragraph{RGM}
The entropy regularization weight is set to $0.05$. For the inner loop, we used the Adam optimizer with learning rate $3 * 10^{-4}$, $\beta_{1} = 0$, $\beta_{2} = 0.999$, and $\epsilon = 10^{-8}$. For the outer loop, we used the Adam optimizer with learning rate $1 * 10^{-4}$, $\beta_{1} = 0$, $\beta_{2} = 0.999$, and $\epsilon = 10^{-8}$.

\paragraph{\notrudder{}}
Apart from the perception module, the same \notrudder{} implementation was used for \texttt{bsuite} as was used for KtD experiments. The actor-critic was trained with the same hyperparameters as the Actor-Critic baseline for \texttt{bsuite}. The LSTM was trained with the same hyperparameters as in KtD.

\section{Atari Experiment}
\subsection{Implementation Details}
Most methods use a similar neural network architecture as described in~\S\ref{subsec:ktd-architecture}. There are two common differences. First, we generate $20$-step trajectories instead of full episodes for each policy update, following the original A2C implementation~\citep{mnih2016asynchronous}. Second, the standard perception module is replaced by the convolutional neural network architecture used in~\citep{mnih2015human}. Further architecture differences and hyperparameters are described below.

\paragraph{A2C}
The policy and the value function share the perception module. The value loss coefficient is $0.5$. The entropy regularization coefficient is $0.01$. We used the RMSProp optimizer with learning rate $0.0007$, decay $0.99$, and $\epsilon = 10^{-5}$. The gradient is clipped to $0.5$ by global norm. The discount factor $\gamma$ is $0.99$. We searched for the eligibility traces parameter $\lambda$ in $\{0.8, 0.9, 0.95, 0.98, 0.99, 1\}$ and selected $0.95$.

\paragraph{Meta-PWTD}
The inner-loop hyperparameters are exactly the same as the A2C baseline. For the outer loop, We applied an entropy regularization as well to stabilize training. The coefficient is $0.01$. The outer loop uses the Adam optimizer with learning rate $0.00003$, $\beta_{1} = 0$, $\beta_{2} = 0.999$, and $\epsilon = 10^{-8}$. The outer-loop gradient is clipped to $0.05$ by global norm.

\paragraph{\notrudder{}}
To handle the variable length episodes in Atari, we chunk the trajectories as described
in the RUDDER paper~\cite{arjona2019rudder}.
Unlike RUDDER, \notrudder{} uses a circular replay buffer, to which all trajectories are added and samples training batches from the buffer uniformly.
We use the quality measure described in the RUDDER paper for mixing the advantage
estimates to the \notrudder{} implementation for Atari.
The quality is computed as described in the paper, and used for mixing the advantage computed with the environment rewards and the one computed with the redistributed rewards.
The quality is also used as the coefficient for the mean-squared error loss used for training the baseline for the redistributed reward.
The LSTM is trained for a maximum of 100 LSTM epochs every 100 actor-critic training iterations.
If the quality of the last 40 LSTM training trajectories is positive after updating the LSTM, the LSTM training is stopped.
For training the LSTM, we normalize the rewards by the maximum return encountered so far and multiply the normalized rewards by $10$.
Before mixing the regular and the redistributed advantages, we denormalize the redistributed rewards by inverting the normalization process above.
The inputs to the LSTM are the delta-frames and one-hot encoded actions.
In the RUDDER paper, a more sophisticated exploration strategy is used for collecting data. We did not implement it for a fair comparison to other methods.

\notrudder{} uses the same A2C agent as the Fixed-$\lambda$ baseline, with the same hyperparameters.
Apart from the differences described above, the LSTM training hyperparameters are from the RUDDER paper~\cite{arjona2019rudder}.
The LSTM is trained with trajectories of length $512$, further split into chunks of length $128$. The LSTM training batch size is 8, learning rate is $10^{-4}$, the optimizer is ADAM with $\epsilon=10^{-8}$, gradient clip is $0.5$, and the $L_2$-regularizer coefficient is $10^{-7}$.
The LSTM starts training after at least $32$ trajectories have been collected from the environment.
The replay buffer stores maximum of $128$ trajectories of length $512$.

\subsection{Additional Empirical Results}
Figure~\ref{fig:atari-learning-curve} shows the learning curves of Meta-PWTD, \notrudder{}, and the A2C baseline.

\begin{figure*}
    \centering
    \subfloat[H-PWR]{\includegraphics[width=0.3\textwidth]{figures/1d_reset/supplementary_weight_matrices/supplementary_handcrafted_8_reward.pdf}}
    \subfloat[Meta-PWR]{\includegraphics[width=0.3\textwidth]{figures/1d_reset/supplementary_weight_matrices/supplementary_8_reward.pdf}}
    \subfloat[Meta-PWTD]{\includegraphics[width=0.3\textwidth]{figures/1d_reset/supplementary_weight_matrices/supplementary_8_td_error.pdf}} \quad
    \subfloat{\includegraphics[width=0.9\textwidth, trim=0 0 10 0]{figures/1d_reset/supplementary_weight_matrices/supplementary_horizontal_colorbar.pdf}}
    \caption{Handcrafted, meta-learned reward weights, and meta-learned TD-error weights in the depth-$8$ DAG environment in \textbf{a}, \textbf{b}, and \textbf{c} respectively. White is unreachable. Y-axis has been cropped to only include weights that affect inner loop learning.}
    \label{fig:supp-depth-8-dag-weights}
\end{figure*}

\begin{figure*}
    \centering
    \subfloat[H-PWR]{\includegraphics[width=\textwidth]{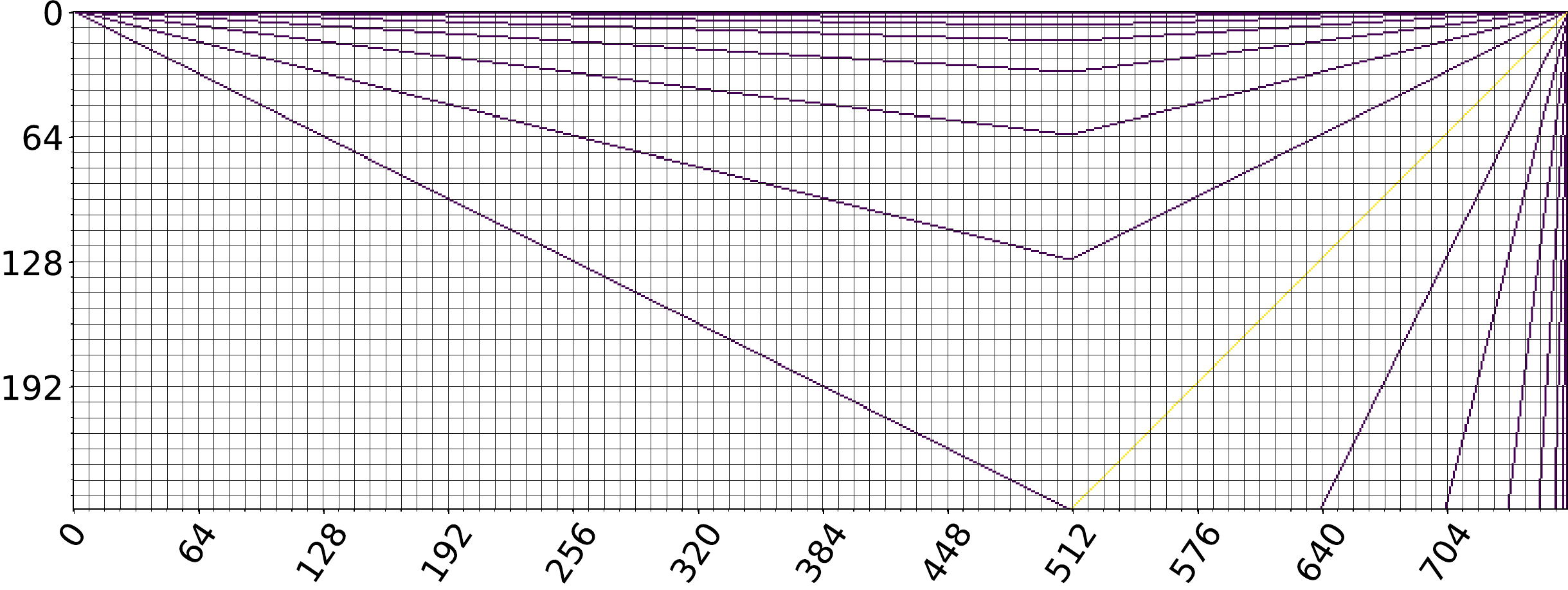}} \quad
    \subfloat[Meta-PWR]{\includegraphics[width=\textwidth]{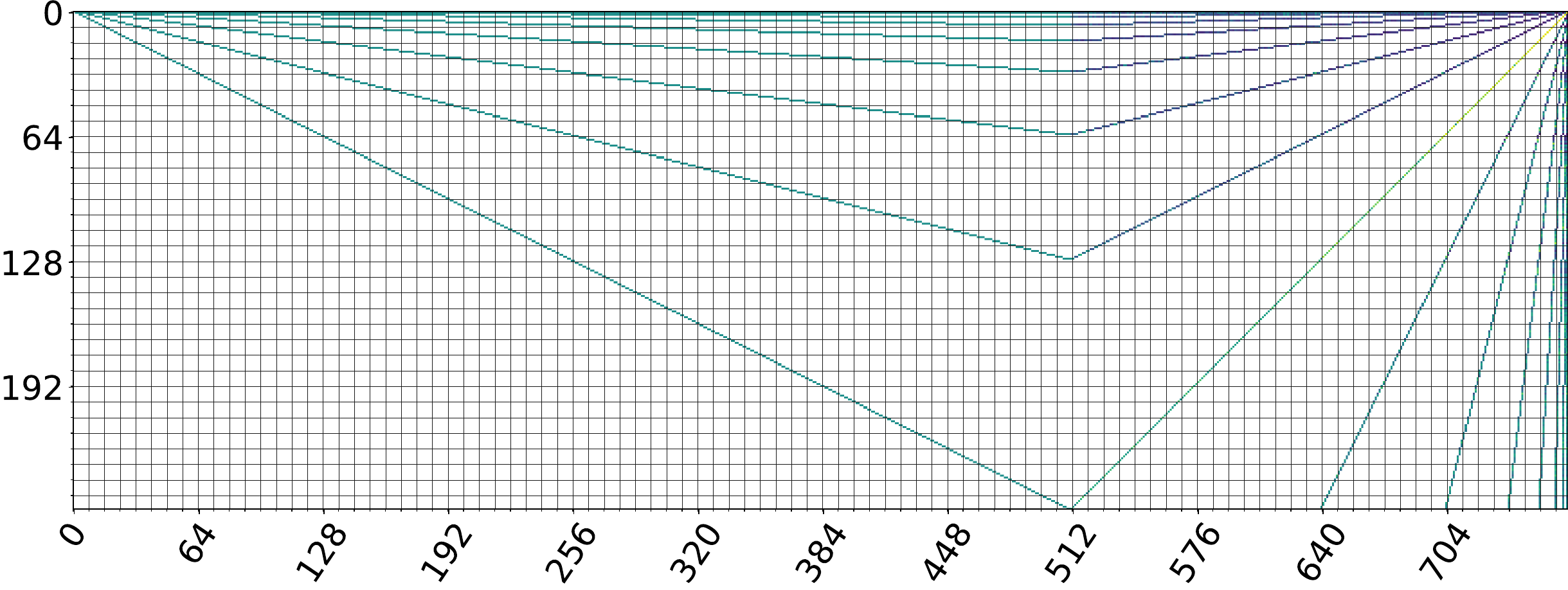}} \quad
    \subfloat[Meta-PWTD]{\includegraphics[width=\textwidth]{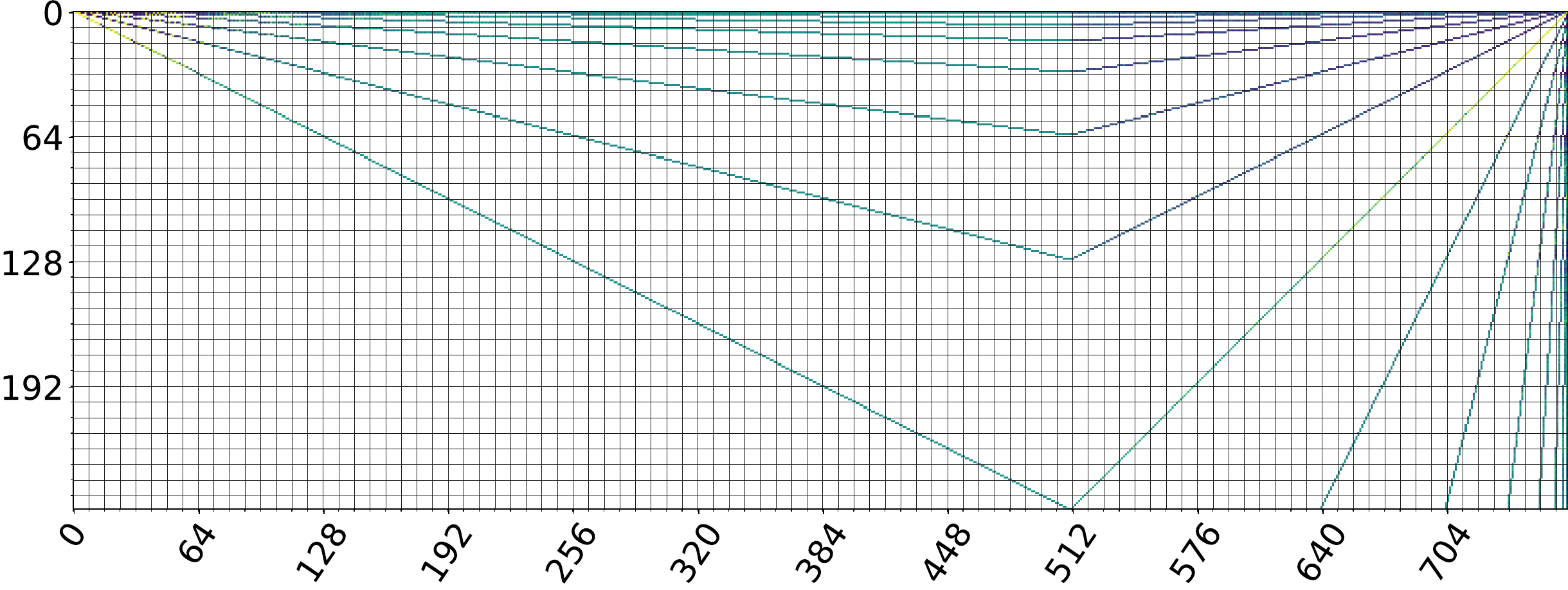}} \quad
    \subfloat{\includegraphics[width=\textwidth, trim=-17 0 8 0]{figures/1d_reset/supplementary_weight_matrices/supplementary_horizontal_colorbar.pdf}}
    \caption{Handcrafted, meta-learned reward weights, and meta-learned TD-error weights in the depth-$16$ DAG environment in \textbf{a}, \textbf{b}, and \textbf{c} respectively. White is unreachable. Y-axis has been cropped to only include weights that affect inner loop learning.}
    \label{fig:supp-depth-16-dag-weights}
\end{figure*}

\begin{figure*}
    \centering
    \subfloat[Mask Depth = 0]{\includegraphics[width=0.3\textwidth]{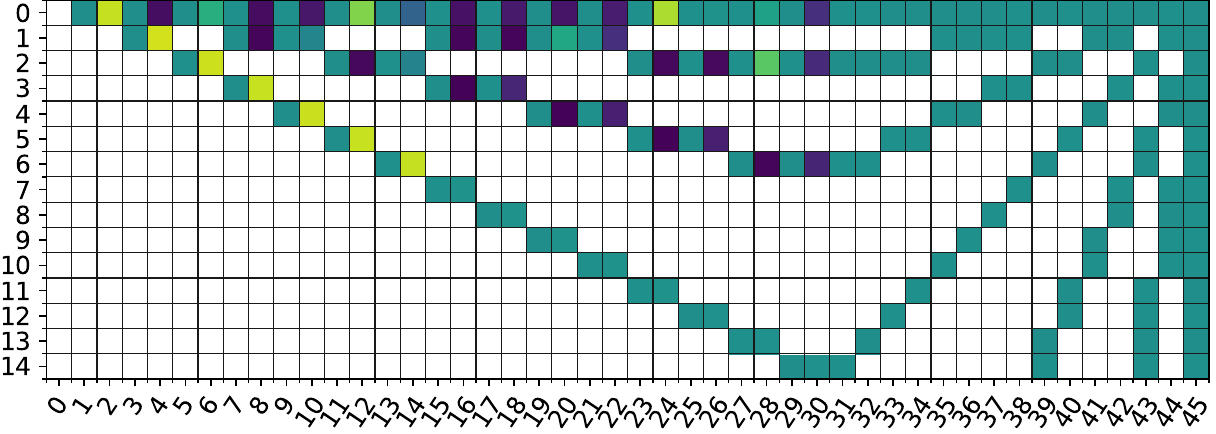}}
    \subfloat[Mask Depth = 4]{\includegraphics[width=0.3\textwidth]{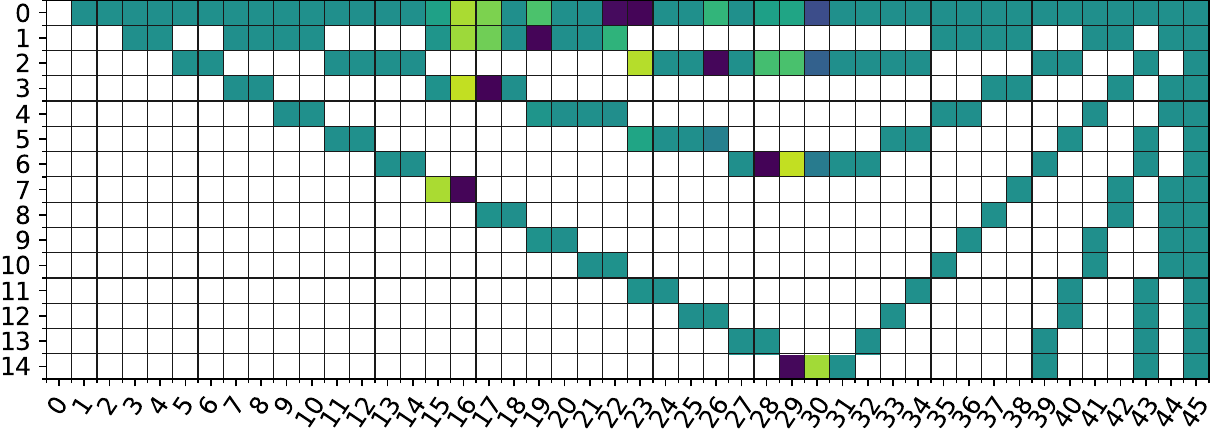}}
    \subfloat[Mask Depth = 8]{\includegraphics[width=0.3\textwidth]{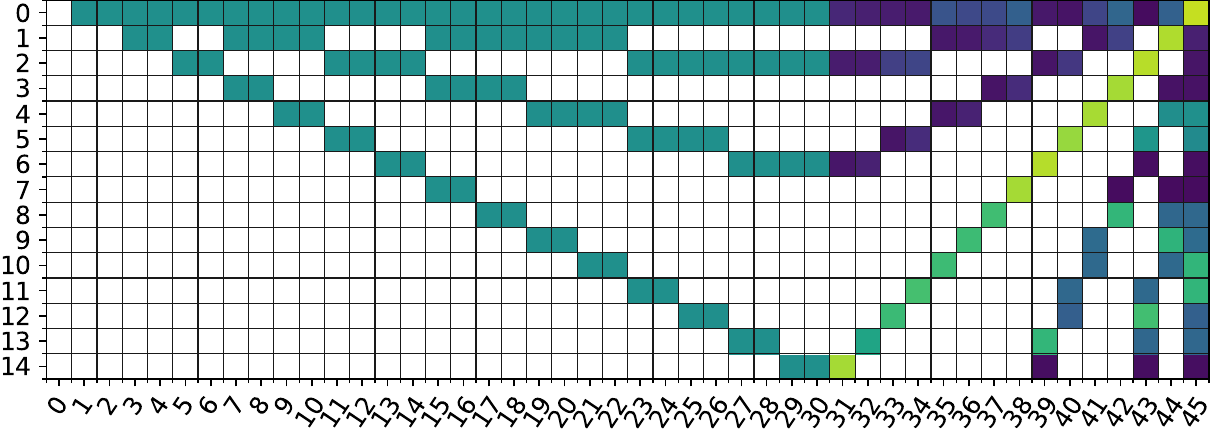}} \quad
    \subfloat{\includegraphics[width=0.5\textwidth, trim=0 0 10 0]{figures/1d_reset/supplementary_weight_matrices/supplementary_horizontal_colorbar.pdf}}
    \caption{Meta-PWTD inner loop-reset weight visualization experiment with masked optimal value function. Figures \textbf{(a,b,c)} show the learned weight matrices for mask depths $0$, $4$, and $8$ respectively.}
    \label{fig:dag_masked_value}
\end{figure*}

\begin{figure*}
  \centering
  {\includegraphics[width=0.9\textwidth]{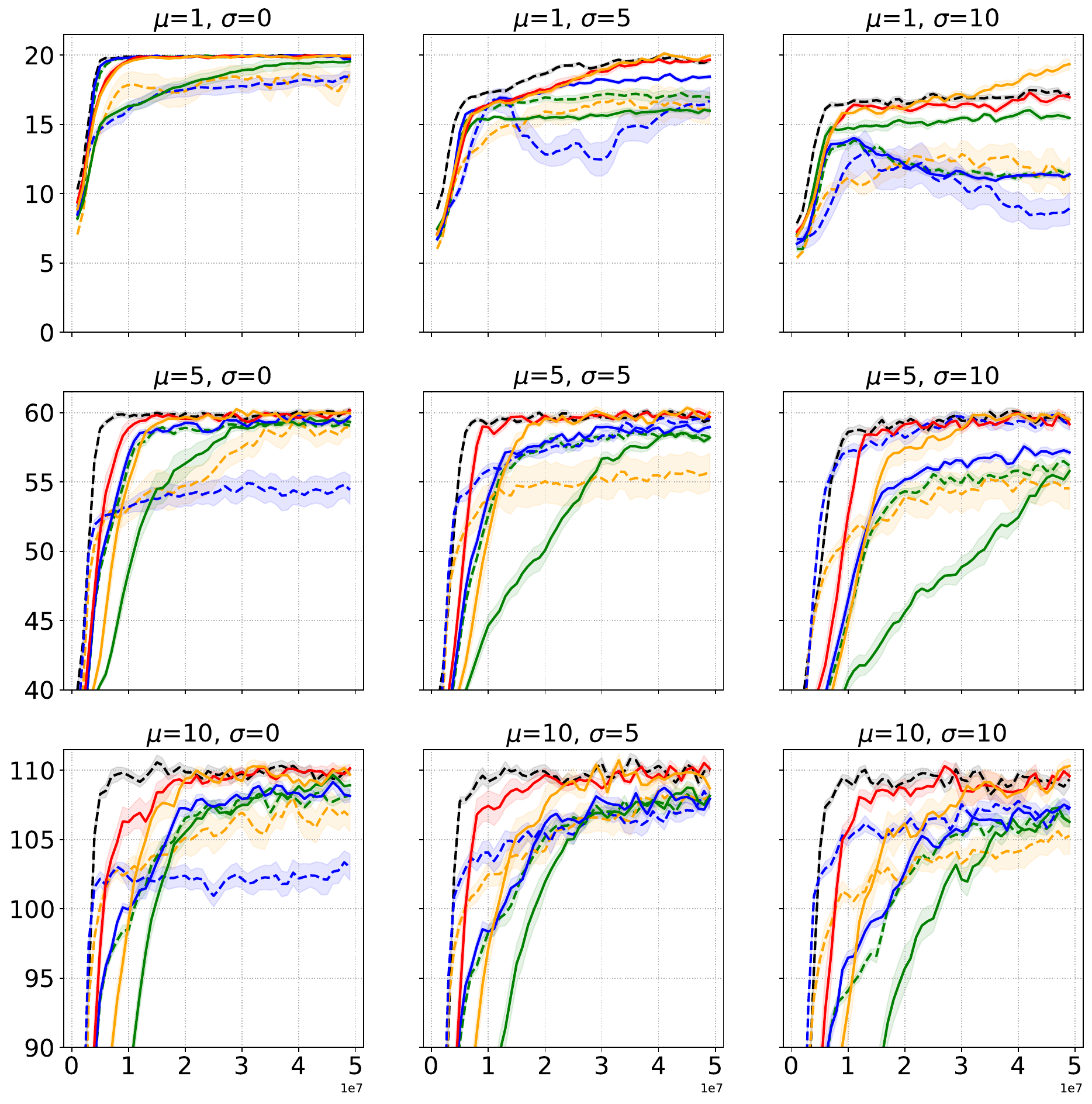}} \quad
  {\includegraphics[width=0.9\textwidth]{figures/key_apple_door/0922-KtD-three-legend.pdf}}
  \caption{Episode returns in all $9$ variants of the Key-to-Door environment. The x-axis is the number of frames. Rows are the different apple reward means ($\mu$), columns the different apple reward variance ($\sigma^{2}$). The x-axis reflects the number of frames, y-axis the episode return. The curves show the average over $10$ independent runs with different random seeds and the shaded area shows the standard errors.}
  \label{fig:ktd-episode-all}
\end{figure*}

\begin{figure*}
  \centering
  {\includegraphics[width=0.9\textwidth]{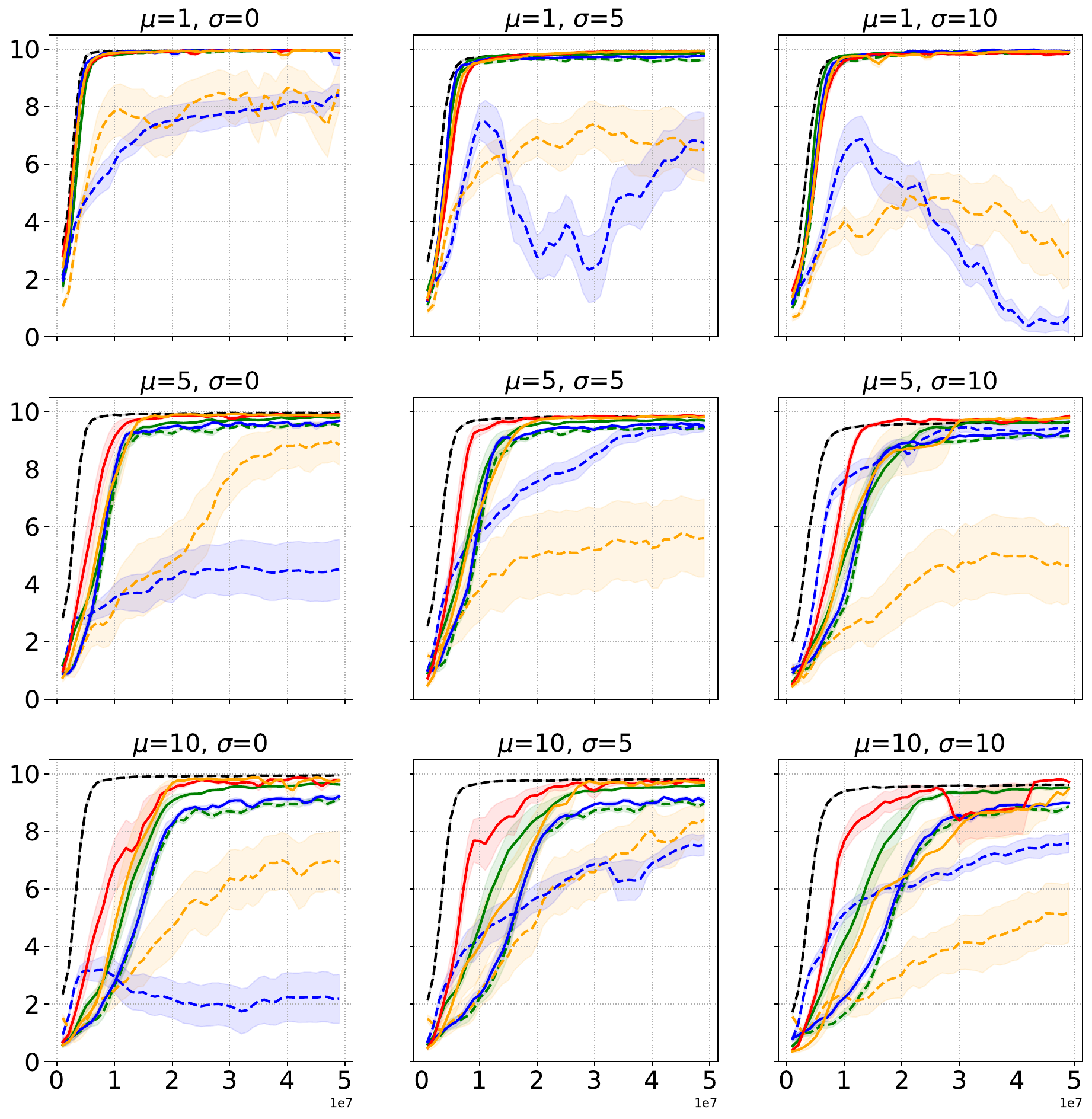}} \quad
  {\includegraphics[width=0.9\textwidth]{figures/key_apple_door/0922-KtD-three-legend.pdf}}
  \caption{Door phase rewards in all $9$ variants of the Key-to-Door environment. The x-axis is the number of frames. Rows are the different apple reward means ($\mu$), columns the different apple reward variance ($\sigma^{2}$). The x-axis reflects the number of frames, y-axis the door phase return. The curves show the average over $10$ independent runs with different random seeds and the shaded area shows the standard errors.}
  \label{fig:ktd-door-all}
\end{figure*}

\begin{figure*}
  \centering
  {\includegraphics[width=0.9\textwidth]{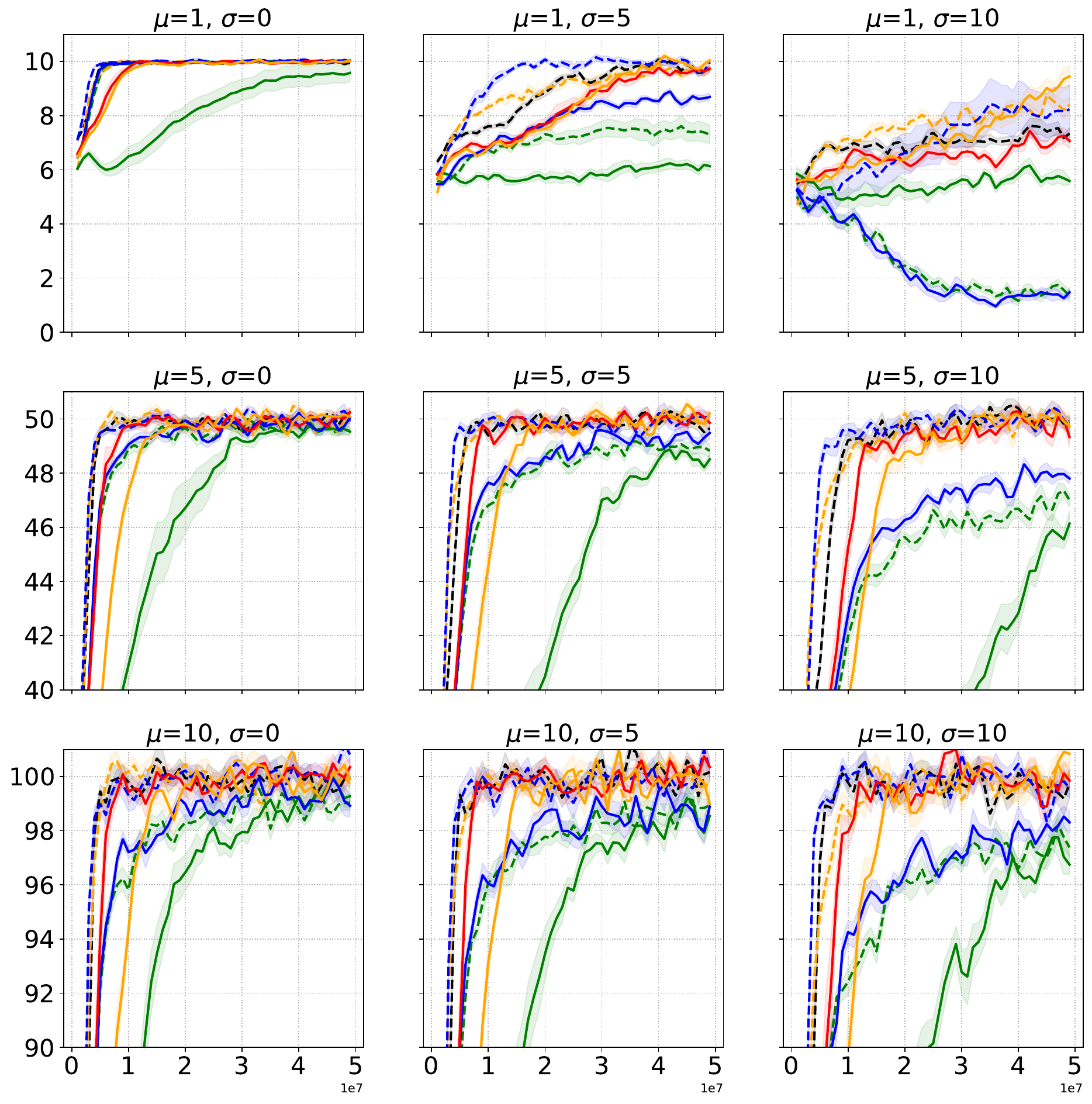}} \quad
  {\includegraphics[width=0.9\textwidth]{figures/key_apple_door/0922-KtD-three-legend.pdf}}
  \caption{Apple phase rewards in all $9$ variants of the Key-to-Door environment. The x-axis is the number of frames. Rows are the different apple reward means ($\mu$), columns the different apple reward variance ($\sigma^{2}$). The x-axis reflects the number of frames, y-axis the apple phase return. The curves show the average over $10$ independent runs with different random seeds and the shaded area shows the standard errors.}
  \label{fig:ktd-apple-all}
\end{figure*}

\begin{figure*}
  \centering
  {\includegraphics[width=0.9\textwidth]{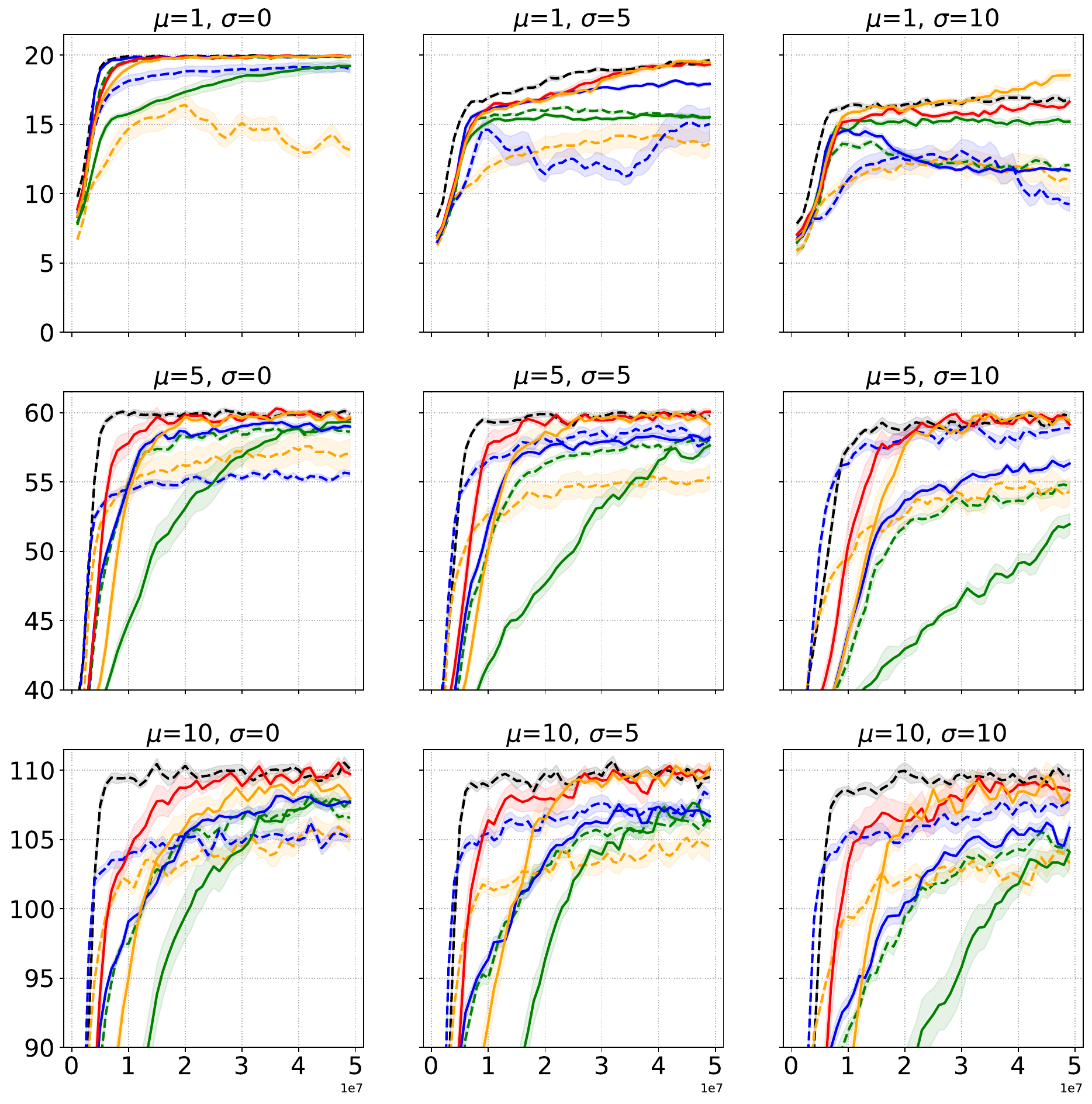}} \quad
  {\includegraphics[width=0.9\textwidth]{figures/key_apple_door/0922-KtD-three-legend.pdf}}
  \caption{Episode returns in all $9$ variants of the stochastic Key-to-Door environment. The x-axis is the number of frames. Rows are the different apple reward means ($\mu$), columns the different apple reward variance ($\sigma^{2}$). The x-axis reflects the number of frames, y-axis the episode return. The curves show the average over $10$ independent runs with different random seeds and the shaded area shows the standard errors.}
  \label{fig:ktd-sto-episode-all}
\end{figure*}

\begin{figure*}
  \centering
  {\includegraphics[width=0.9\textwidth]{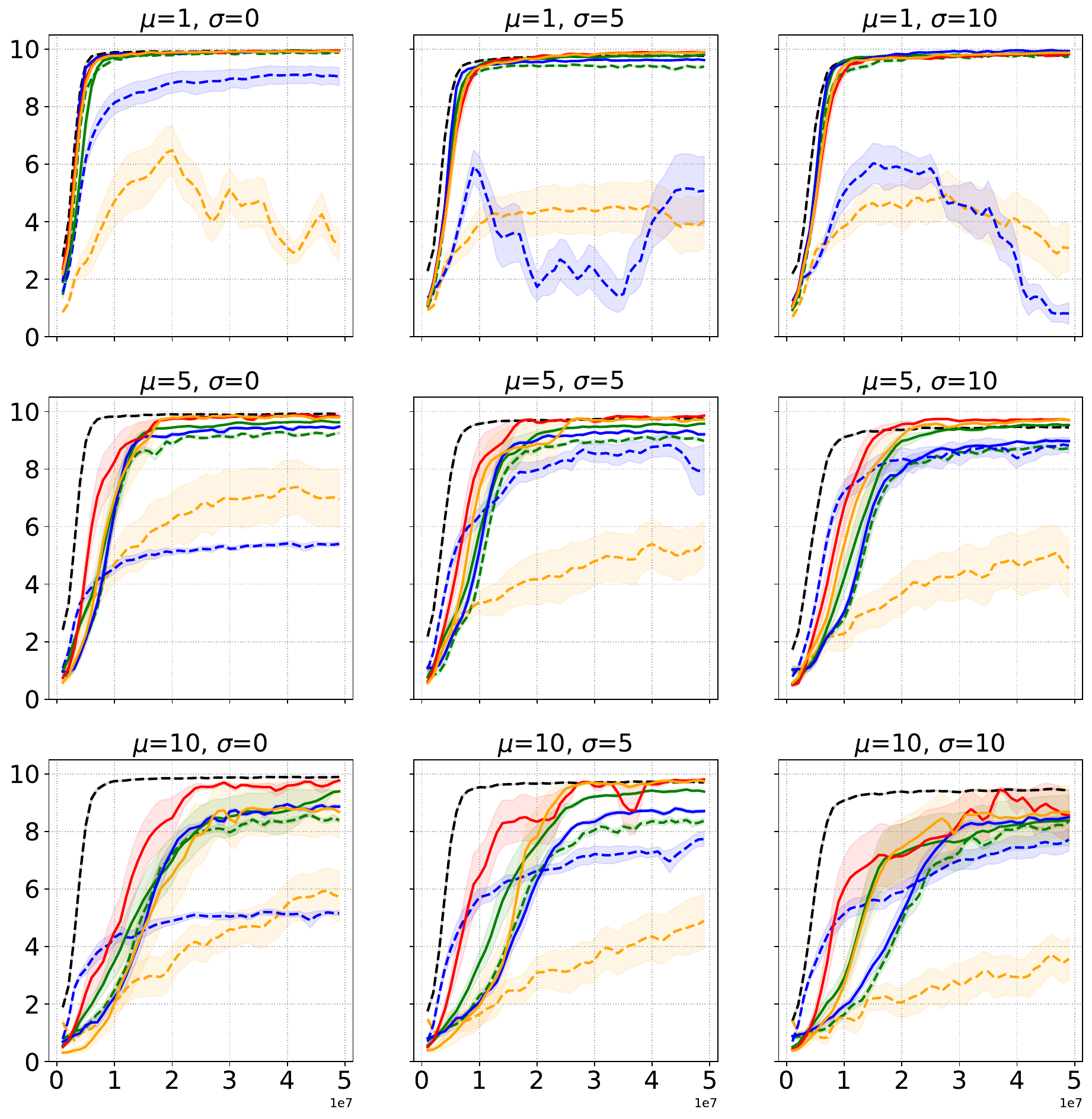}} \quad
  {\includegraphics[width=0.9\textwidth]{figures/key_apple_door/0922-KtD-three-legend.pdf}}
  \caption{Door phase rewards in all $9$ variants of the stochastic Key-to-Door environment. The x-axis is the number of frames. Rows are the different apple reward means ($\mu$), columns the different apple reward variance ($\sigma^{2}$). The x-axis reflects the number of frames, y-axis the door phase return. The curves show the average over $10$ independent runs with different random seeds and the shaded area shows the standard errors.}
  \label{fig:ktd-sto-door-all}
\end{figure*}

\begin{figure*}
  \centering
  {\includegraphics[width=0.9\textwidth]{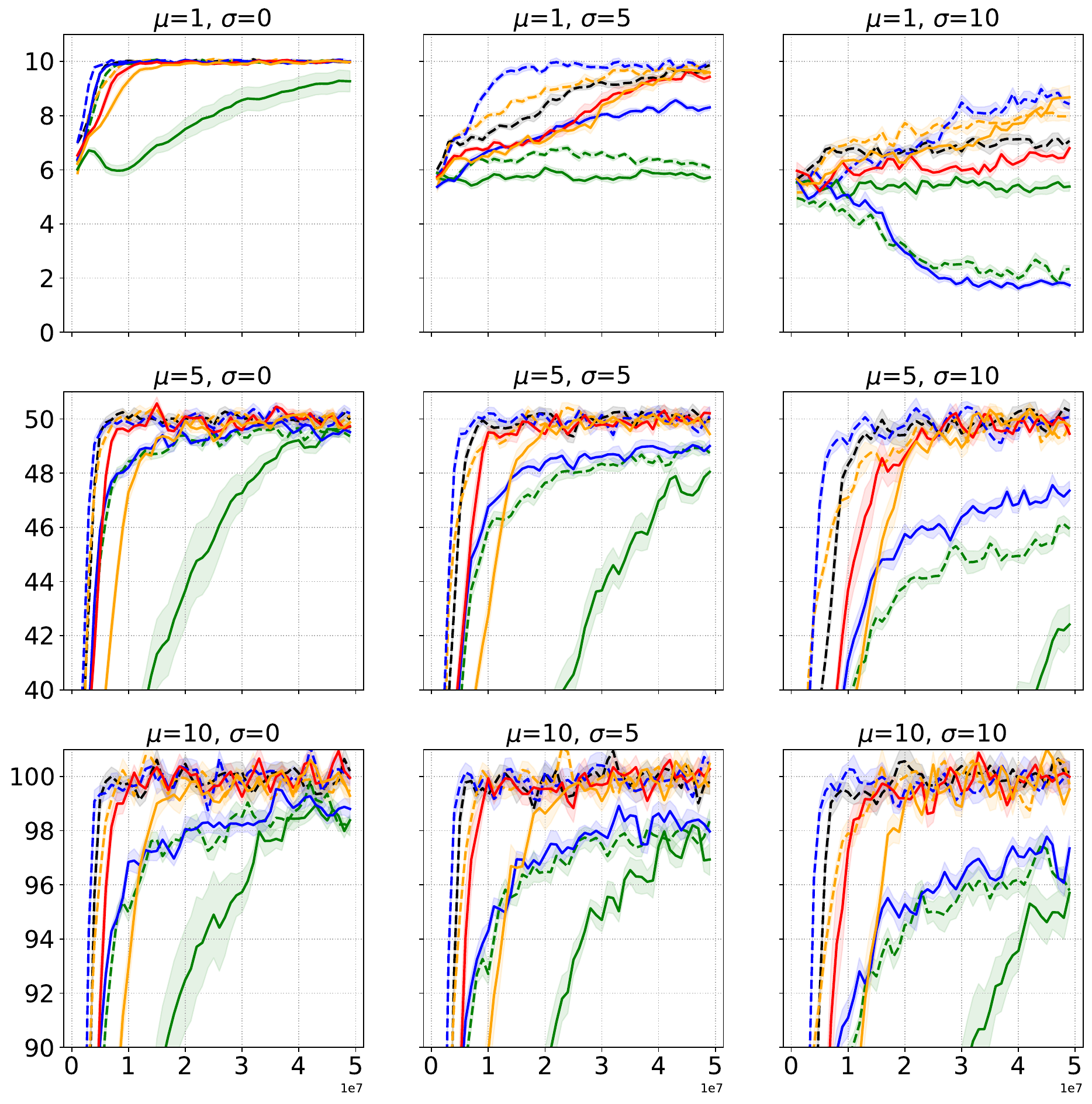}} \quad
  {\includegraphics[width=0.9\textwidth]{figures/key_apple_door/0922-KtD-three-legend.pdf}}
  \caption{Apple phase rewards in all $9$ variants of the stochastic Key-to-Door environment. The x-axis is the number of frames. Rows are the different apple reward means ($\mu$), columns the different apple reward variance ($\sigma^{2}$). The x-axis reflects the number of frames, y-axis the apple phase return. The curves show the average over $10$ independent runs with different random seeds and the shaded area shows the standard errors.}
  \label{fig:ktd-sto-apple-all}
\end{figure*}

\begin{figure*}
  \centering
  {\includegraphics[width=0.85\textwidth]{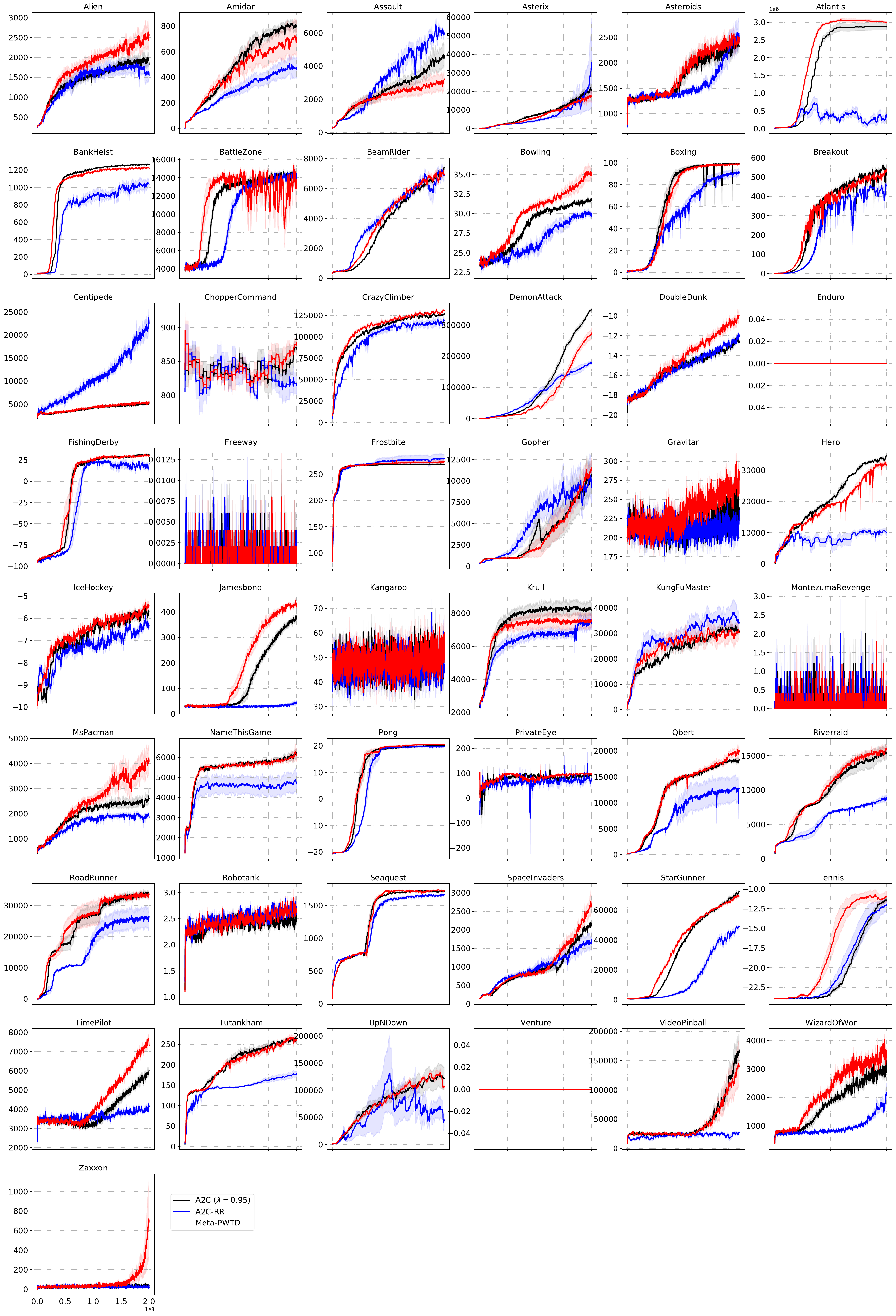}}
  \caption{Learning curves on $49$ Atari games. The x-axis is the number of frames and the y-axis is the episode return. Each curve is averaged over $5$ independent runs with different random seeds. Shaded area shows the standard error over $5$ runs.}
  \label{fig:atari-learning-curve}
\end{figure*}

\end{document}